\documentclass[utf8]{article} 

\usepackage{arxiv}

\usepackage{url,hyperref,microtype,subcaption}
\usepackage[onehalfspacing]{setspace}

\usepackage{mathtools}
\usepackage{amsmath,amssymb,amsfonts}
\usepackage[sort&compress,nameinlink,capitalize]{cleveref}
\usepackage{algorithmic}
\usepackage{graphicx}
\usepackage{textcomp}
\usepackage{xcolor}
\usepackage{rotating} 
\usepackage{tikz}
\usetikzlibrary{trees, positioning, arrows.meta}

\usepackage{adjustbox}
\usepackage{pifont}
\usepackage{colortbl}
\usepackage{booktabs}
\usepackage{multirow}
\usepackage{natbib}

\crefname{paragraph}{Section}{Sections}
\Crefname{paragraph}{Section}{Sections}

\renewcommand{\arraystretch}{1.1}

\newcommand{\bxmark}{\ding{55}}
\newcommand{\bcmark}{\ding{51}}
\newcolumntype{L}[1]{>{\raggedright\arraybackslash}p{#1}}
\newcolumntype{C}[1]{>{\centering\arraybackslash}p{#1}}
\definecolor{lightgray}{gray}{0.6}

\title{Diffusion Models for Robotic Manipulation: A Survey}

\author{
 Rosa Wolf \\
  Karlsruhe Institute of Technology (KIT)\\
  Karlsruhe, Germany \\
  \texttt{rosa.wolf@kit.edu} \\
   \And
 Yitian Shi \\
   Karlsruhe Institute of Technology (KIT)\\
  Karlsruhe, Germany \\
  \And
 Sheng Liu \\
   Karlsruhe Institute of Technology (KIT)\\
  Karlsruhe, Germany \\
  \And
 Rania Rayyes \\
   Karlsruhe Institute of Technology (KIT)\\
  Karlsruhe, Germany \\
}

\begin{document}
\maketitle

\begin{abstract}
Diffusion generative models have demonstrated remarkable success in visual domains such as image and video generation. They have also recently emerged as a promising approach in robotics, especially in robot manipulations. Diffusion models leverage a probabilistic framework, and they stand out with their ability to model multi-modal distributions and their robustness to high-dimensional input and output spaces. 
This survey provides a comprehensive review of state-of-the-art diffusion models in robotic manipulation, including grasp learning, trajectory planning, and data augmentation. Diffusion models for scene and image augmentation lie at the intersection of robotics and computer vision for vision-based tasks to enhance generalizability and data scarcity. This paper also presents the two main frameworks of diffusion models and their integration with imitation learning and reinforcement learning. In addition, it discusses the common architectures and benchmarks and points out the challenges and advantages of current state-of-the-art diffusion-based methods.

\end{abstract}

\section{Introduction}

Diffusion Models (DMs) have emerged as highly promising deep generative models in diverse domains, including computer vision \citep{Ho2020DenoisingModels, Song2021DenoisingModels, Nichol2021ImprovedModels, Ramesh2022HierarchicalLatents, Rombach2021High-ResolutionModels}, natural language processing \citep{NEURIPS2022_1be5bc25, NEURIPS2023_fdba5e0a, pmlr-v162-yu22h}, and robotics \citep{Chi2023DiffusionDiffusion, Urain2023SE3-DiffusionFields:Diffusion}. 
 DMs intrinsically posses the ability to model any distribution. They have demonstrated remarkable performance and stability in modeling complex and multi-modal distributions\footnote{In the context of probability distributions, ``multi-modal'' does not refer to multiple input modalities but rather to the presence of multiple peaks (modes) in the distribution, each representing a distinct possible outcome. For example, in trajectory planning, a multi-modal distribution can capture multiple feasible trajectories. Accurately modeling all modes is crucial for policies, as it enables better generalization to diverse scenarios during inference.} from high-dimensional and visual data surpassing the ability of Gaussian Mixture Models (GMMs) or Energy-based models (EBMs) like Implicit behavior cloning (IBC) \citep{Chi2023DiffusionDiffusion}. While GMMs and IBCs can model multi-modal distributions, and IBCs can even learn complex discontinuous distributions \citep{Florence2022ImplicitCloning}, experiments \citep{Chi2023DiffusionDiffusion} show that in practice, they might be heavily biased toward specific modes. In general, DMs have also demonstrated performance exceeding generative adversarial networks (GANs) \citep{Krichen2023GenerativeNetworks}, which were previously considered the leading paradigm in the field of generative models. GANs usually require adversarial training, which can lead to mode collapse and training instability \citep{Krichen2023GenerativeNetworks}. 
 Additionally, GANs have been reported to be sensitive to hyperparameters \citep{Lucic2018AreStudy}.

Since 2022, there has been a noticeable increase in the implementation of diffusion probabilistic models within the field of robotic manipulation. These models are applied across various tasks, including trajectory planning, e.g., \citep{Chi2023DiffusionDiffusion} and grasp prediction, e.g., \citep{Urain2023SE3-DiffusionFields:Diffusion}. The ability of DMs to model multi-modal distributions is a great advantage in many robotic manipulation applications. In various manipulation tasks, such as trajectory planning and grasping, there exist multiple equally valid solutions (redundant solutions). Capturing all solutions improves generalizability and robots' versatility, as it enables generating feasible solutions under different conditions, such as different placements of objects or different constraints during inference. Although in the context of trajectory planning using DMs, primarily imitation learning is applied, DMs have been adapted for integration with reinforcement learning (RL), e.g., \citep{Geng2023DiffusionStrategies}. Research efforts focus on various components of the diffusion process adapted to different tasks in the domain of robotic manipulation.
To give just some examples, developed architectures integrate different or even multiple input modalities. One example of an input modality could be point clouds \citep{Ze20243DRepresentations, Ke20243DRepresentations}. With the provided depth information, models can learn more complex tasks, for which a better 3D scene understanding is crucial. Another example of an additional input modality could be natural language \citep{Ke20243DRepresentations, Du2023LearningGeneration, Li2024GeneralizableGuidance}, which also enables the integration of foundation models, like large language models, into the workflow. In \cite{Ze20243DRepresentations}, both point clouds and language task instructions are used as multiple input modalities.
Others integrate DMs into hierarchical planning \citep{Ma2024HierarchicalManipulation,Du2023LearningGeneration} or skill learning \citep{Liang2023SkillDiffuser:Execution, Mishra2023GenerativeModels}, to facilitate their state-of-the-art capabilities in modeling high-dimensional data and multi-modal distributions, for long-horizon and multi-task settings.
Many methodologies, e.g. \citep{Kasahara2023RIC:Reconstruction, Chen2023GenAug:Environments}, employ diffusion-based data augmentation in vision-based manipulation tasks to scale up datasets and reconstruct scenes. It is important to note that one of the major challenges of DMs is its comparatively slow sampling process, which has been addressed in many methods, e.g., \citep{Song2021DenoisingModels,Chen2024DontDiffusion, Zhou2024VariationalExperts}, also enabling real-time prediction.

To the best of our knowledge, we provide the first survey of DMs concentrating on the field of robotic manipulation. The survey offers a systematic classification of various methodologies related to DMs within the realm of robotic manipulation, regarding network architecture, learning framework, application, and evaluation. Alongside comprehensive descriptions, we present illustrative taxonomies.

To provide the reader with the necessary background information on DMs, we will first introduce their fundamental mathematical concepts (\cref{chap:math}). This section provides a general overview of DMs rather than focusing specifically on robotic manipulation.
Then, network architectures commonly used for DMs in robotic manipulation will be discussed (\cref{chap:architecture}).
Next (\cref{chap:applications}), we explore the three primary applications of DMs in robotic manipulation: trajectory generation (\cref{chap:traj-planning}), robotic grasp synthesis (\cref{chap:grasp-learning}), and visual data augmentation (\cref{chap:augmentation}). This is followed by an overview of commonly used benchmarks and baselines (\cref{chap:benchmarks}). Finally, we discuss our conclusions and existing limitations, and outline potential directions for future research (\cref{chap:conclusion}).

\section{Preliminaries on Diffusion Models}\label{chap:math}

\begin{figure}[ht!]

\includegraphics[width=.96\linewidth]{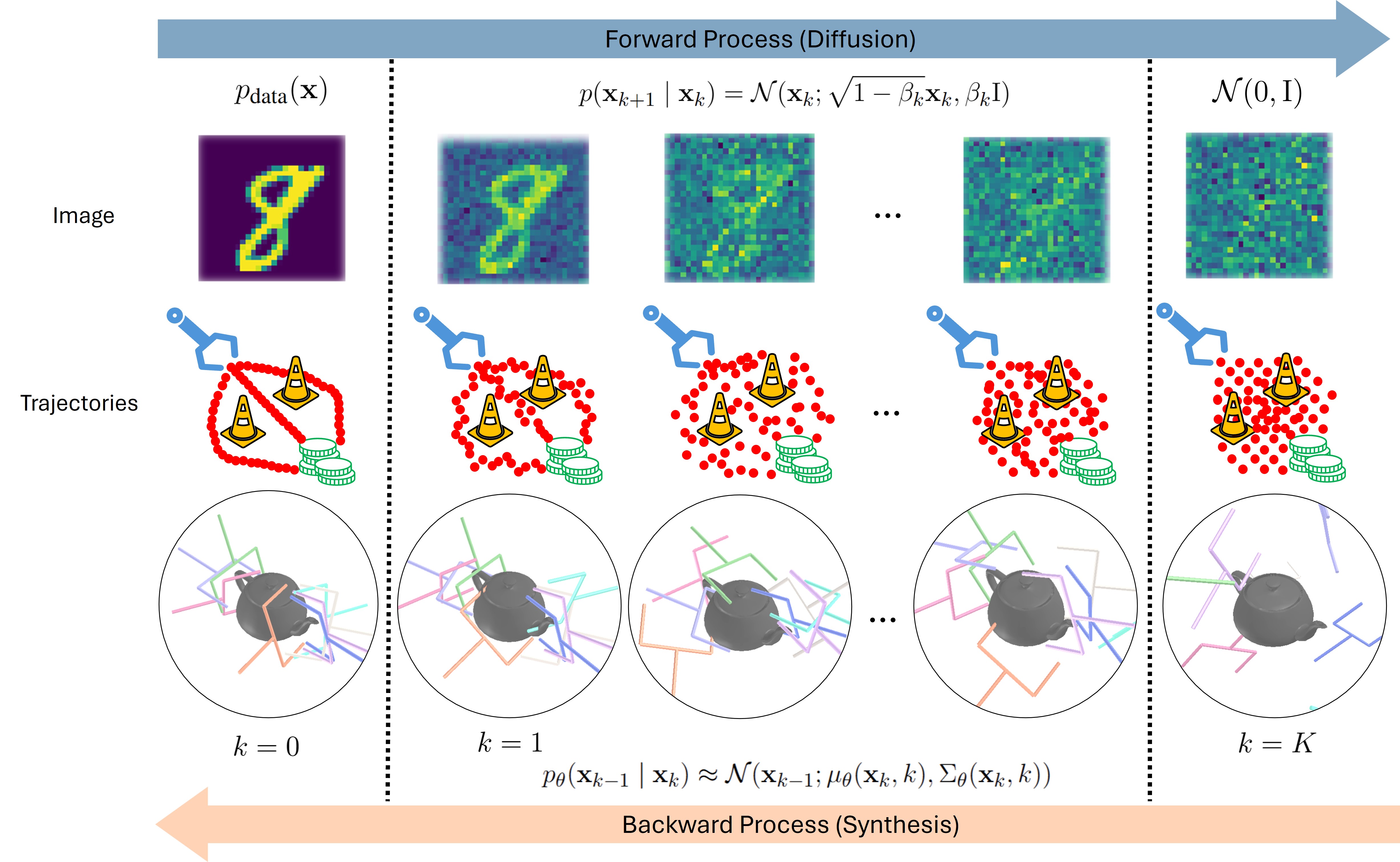}
    \caption{\small Illustrations of diffusion (forward) processes on image, trajectories, and grasp poses~(\cite{Urain2023SE3-DiffusionFields:Diffusion}) and their corresponding synthesis (backward) processes.}
    \label{fig:fig1}
\end{figure}

\subsection{Mathematical Framework}

The key idea of DMs is to gradually perturb an unknown target distribution~$p_\text{data}(x)$ into a simple known distribution, e.g., a normal Gaussian distribution, which is first introduced in \citep{Sohl-Dickstein2015DeepThermodynamics}. To generate new data, points are sampled from the initial known “simple” distribution, and perturbations are estimated to iteratively reverse the diffusion process. The forward and backward diffusion processes are also visualized in Fig. \ref{fig:fig1}. There exist two main approaches to diffusion-based modeling, both based on the original work by \cite{Sohl-Dickstein2015DeepThermodynamics}. The first group of methods is score-based DMs, where the gradient of the log-likelihood of the data is learned to reverse the diffusion process. This score-based generative modeling was first introduced in \cite{Song2019GenerativeDistribution}. In the other group of methods, a network is trained to directly predict the noise, which is added during the forward process. This methodology was first introduced in Denoising Diffusion Probabilistic Models (DDPM) \citep{Ho2020DenoisingModels}.

The original score-based DM by \cite{Song2019GenerativeDistribution} is rarely used in the field of robotic manipulation. This could be due to its inefficient sampling process. However, as it forms a crucial mathematical framework and baseline for many of the later developed DMs, e.g. \citep{Song2021Score-BasedEquations, Karras2022ElucidatingModels}, including DDPM \cite{Ho2020DenoisingModels}, we describe the main concepts in the following section. While DDPM is rarely used as well, the commonly used method Denoising Diffusion Implicit Models (DDIM) \citep{Song2021DenoisingModels} originates from DDPM. DDIM only alters the sampling process of DDPM while keeping its training procedure. Hence, understanding DDPM is crucial for many applications of DMs in robotic manipulation.

In the following sections, we first introduce score-based DMs, then DDPM, before addressing their shortcomings.

\subsubsection{Denoising Score Matching using Noise Conditional Score Networks}\label{chap:NCSM}
One approach to estimate perturbations in the data distribution is to use denoising score matching with Lagenvin dynamics (SMLD), where the score of the data density of the perturbed distributions is learned using a Noise Conditional Score Network (NCSM)\citep{Song2019GenerativeDistribution}. This method is described in this section, and for more details, please refer to their original work.
During the forward diffusion process, data~$\textbf{x}$ from an unknown distribution~$p_\text{data}(\textbf{x})$ is transformed into random noise~$\mathcal{N}(0, \mathrm{I})$, by gradually adding noise. New data is generated during the reverse process, where the learned NCSM is used to iteratively denoise the initial samples.

\paragraph{Forward Process}
Let~$\{\sigma_k\}_{k = 1}^K$ be a noise schedule with progressively increasing variance, i.e., $\sigma_k < \sigma_{k + 1}$ for all~$k \in \{1, \dots, K\}$.
To get from the true data distribution~$p_\text{data}(\mathbf{x})$ to the perturbed data distribution~$p_{\sigma_k}(\mathbf{x}_k)$, with variance~$\sigma_k$, noise is added to the data according to a pre-specified noise distribution~$p_{\sigma_k}(\mathbf{x}_k \mid \mathbf{x})$. 
To denoise the data, the gradients of the logarithmic probability density functions $\nabla_\mathbf{x} \log p_{\sigma_k}(\mathbf{x}_k \mid \mathbf{x})$, i.e., the scores, are estimated using the NCSM.
To train the NCSM~$\mathbf{s}_\theta(\mathbf{x}_k, \sigma_k)$, for all noise scales~$k \in \{1, \dots, K\}$ the weighted sum of denoising score matchings is minimized~\citep{Song2019GenerativeDistribution}:
\begin{equation}\label{eq:denoising-score-matching}
        \mathcal{L} = \frac{1}{2K}\sum_{k=1}^K \sigma_k^2 \mathbb{E}_{p_\text{data}(\mathbf{x})}\mathbb{E}_{\mathbf{x}_k \sim p_{\sigma_k}(\mathbf{x}_k \mid \mathbf{x})} \left[\left|\left| \nabla_{\mathbf{x}_k} p_{\sigma_k}(\mathbf{x}_k \mid \mathbf{x}) - \mathbf{s}_\theta(\mathbf{x}_k, \sigma_k) \right| \right|_2^2\right] .
\end{equation}

\paragraph{Reverse Process}
Starting with randomly drawn noise samples~$x_K^0 \in \mathcal{N}(0, I)$, Langevin dynamics are applied recursively over all~$k \in \{0, ..., K\}$, to generate samples using the learned score function:
\begin{equation}\label{eq:Langevin-dynamics}
    \mathbf{x}_k^{n} = \mathbf{x}_k^{n - 1} + \alpha_k\mathbf{s}_\theta(x_k^{n - 1}, \sigma_k) + \sqrt{2\alpha_k}\mathbf{z}_k^n, \quad n \in \{0, .., N\},
\end{equation}
where~$\alpha_k > 0$ is the step size and~$z_k^n \in \mathcal{N}(0, I)$ is randomly drawn noise.  During one Langevin dynamic for noise scale~$k$, the index~$n$ is increasing until~$n=N$.
Then, the final value~$\mathbf{x}_k^N$, of one Langevin dynamic becomes the the initial value~$\mathbf{x}_{k - 1}^0$ for the next Langevin dynamic with the next lower noise scale~$k-1$, i.e., $x_{k-1}^0 = x_k^N$. For small enough step sizes, the final generated samples~$\mathbf{x}_0^N$, should be approximately distributed according to~$p_\text{data}(\mathbf{x})$.

\subsubsection{Denoising Diffusion Probabilistic Models (DDPM)}\label{chap:DDPM}

In DDPM \citep{Ho2020DenoisingModels}, instead of estimating the score function directly, a noise prediction network, conditioned on the noise scale, is trained. Similarly to SMLD with NCSN, new points are generated by sampling Gaussian noise and iteratively denoising the samples using the learned noise prediction network.

Notably, there is one step per noise scale in the denoising process instead of recursively sampling from each noise scale. 

\paragraph{Forward Process} To train the noise prediction network~$\epsilon_\theta$, first points~$\mathbf{x_0} \sim p_\text{data}(\mathbf{x})$ are sampled from the true unknown data distribution. The samples are degraded by adding noise~$\epsilon \in \mathcal{N}(0, \mathrm{I})$ until at degrading step~$K$, the degraded samples are approximately normally distributed, i.e. $x_K \sim \mathcal{N}(0, \mathrm{I})$.
As already introduced by \cite{Sohl-Dickstein2015DeepThermodynamics}, the noise is added according to a Markovian process:
\begin{equation}
    p(\mathbf{x}_{k+1} \mid \mathbf{x}_k) = \mathcal{N}(\mathbf{x}_k; \sqrt{1- \beta_k} \mathbf{x}_k, \beta_k \mathrm{I}),
\end{equation}
where~$\beta_1, ..., \beta_K \in [0, 1)$ is the noise variance schedule, which can either be a hyperparameter \citep{Ho2020DenoisingModels}, or optimized as part of the model training process \citep{Nichol2021ImprovedModels}. In practice, instead of adding noise iteratively, the formulation also allows adding the noise in closed form:
\begin{equation}\label{eq:noising-closed-form}
    p(\mathbf{x}_{k+1} \mid \mathbf{x}_0) = \mathcal{N}(\mathbf{x}_k; \sqrt{\bar\alpha_k}\mathbf{x}_0, (1-\bar\alpha_k)\mathrm{I}),
\end{equation}
with~$\bar\alpha_k \coloneq \prod_{i=1}^k (\alpha_i)$ and~$\alpha_k \coloneq 1 - \beta_k$. This allows first uniformly sampling a noise scale~$k \sim \mathcal{U}\{1, K\}$, and then directly inferring the corresponding degraded sample.

Adding the noise in closed form facilitates training a noise prediction network~$\epsilon_\theta(\mathbf{x}_k, k)$ by minimizing the mean squared error for~$k \in \{1, ..., K\}$:
\begin{equation}\label{eq:DDPM-loss}
    \mathcal{L} = \mathbb{E}_{k, \mathbf{x}_0, \epsilon}\left[\left|\left| \epsilon - \epsilon_\theta( \mathbf{x}_k, k)\right|\right|_2^2\right].
\end{equation}


\paragraph{Reverse Process}
Similar to the reverse process described in \cref{chap:NCSM}, new samples are generated from random noise~$\mathbf{x}_K \sim \mathcal{N}(0, \mathrm{I})$, using the learned forward process~$p(\mathbf{x}_k \mid \mathbf{x}_{k - 1})$. 
As the forward process is modeled using Gaussian distributions, the reverse process~$p_\theta(\mathbf{x}_{k - 1} \mid \mathbf{x}_k)$ is also a Gaussian distribution if the number of diffusion steps is sufficiently large, i.e the step size is small enough \citep{Sohl-Dickstein2015DeepThermodynamics}:
\begin{equation}
    p_\theta(\mathbf{x}_{k - 1} \mid \mathbf{x}_k) \approx \mathcal{N}(\mathbf{x}_{k - 1}; \mu_\theta(\mathbf{x}_k, k), \Sigma_\theta(\mathbf{x}_k, k)).
\end{equation}
In DDPM, the variance-schedule is fixed and thus~$\Sigma_\theta(x_k, k) = \beta_k \mathrm{I}$.
Additionally, using reparameterization, it can be shown that the mean of the distribution at each step can be iteratively predicted using the previous value~$\mathbf{x}_k$ and the estimated noise~$\epsilon_\theta$ \citep{Ho2020DenoisingModels}:
\begin{equation}\label{eq:denoising-process}
    \mathbf{x}_{k - 1} = \frac{1}{\sqrt{\alpha_k}}\left(\mathbf{x}_k - \frac{1 - \alpha_k}{\sqrt{1 - \bar\alpha_k}}\epsilon_\theta(\mathbf{x}_k, k)\right) + \sigma_k \mathbf{z}, 
\end{equation}
which is repeated until~$\mathbf{x}_0$ is computed. As in SMLD, for small enough step sizes, the final generated samples~$\mathbf{x}_0$ are approximately distributed according to the true data distribution~$p_\text{data}(x)$.

\subsection{Architectural Improvements and Adaptations}\label{chap:arch-improv-and-adapt}

One of the main disadvantages of DMs is the iterative sampling, leading to a relatively slow sampling process. In comparison, using GANs or variational autoencoders (VAEs), only a single forward pass through the trained network is required to produce a sample. In both DDPM and the original formulation of SMLD, the number of time steps (noise levels) in the forward and reverse processes is equal. While reducing the number of noise levels leads to a faster sampling process, it comes at the cost of sample quality.
Thus, there have been numerous works to adapt the architectures and sampling processes of DDPM and SMLD to improve both the sampling speed and quality of DMs, e.g., \citep{Nichol2021ImprovedModels, Song2021DenoisingModels, Song2021Score-BasedEquations}.

\subsubsection{Improving Sampling Speed and Quality}\label{chap:sampling-speed}

The forward diffusion process can be formulated as a stochastic differential equation (SDE). Using the corresponding reverse-time SDE, SDE-solvers can then be applied to generate new samples \citep{Song2021Score-BasedEquations}. \cite{Song2021Score-BasedEquations} shows that the diffusion process from SMLD corresponds to an SDE where the variance of the perturbation kernels~$\{p(x_k \mid x_0)\}_{k=1}^K$ is exploding with increasing~$K$. This is referred to as the variance exploding SDE (VE SDE) in the literature. The diffusion process from DDPM corresponds to a variance-preserving SDE, referred to as VP SDE in the literature.
As such, the original formulations of SMLD and DDPM can be interpreted as specific discretizations of their corresponding SDEs.
\cite{Song2021Score-BasedEquations} also shows that once the score-network is trained, the reverse-time SDE can be replaced by an ordinary differential equation (ODE). Using an ODE has several advantages. As the reverse process is deterministic, it allows for precise likelihood computation \citep{Song2021Score-BasedEquations}. Moreover, the deterministic process naturally leads to higher consistency. Thus, the ODE formulation can be used as a high-level feature-preserving encoding, which also allows interpolations in latent space \citep{Song2021DenoisingModels, Karras2022ElucidatingModels}. Finally, using ODEs enables faster and adaptive sampling, which is why it forms the baseline for many of the following methods.


One group of methods aimed at improving sampling speed \citep{Jolicoeur-Martineau2021GottaModels, Song2021DenoisingModels, Lu2022DPM-Solver:Steps, Karras2022ElucidatingModels} designs samplers that operate independently of the specific training process. Using an SDE/ODE-based formulation allows choosing different discretizations of the reverse process than for the forward process. Larger step sizes reduce computational cost and sampling time but introduce greater truncation error. The sampler operates independently of the specific noise prediction network implementation, enabling the use of a single network, such as one trained with DDPM, with different samplers.

Denoising Diffusion Implicit Models (DDIM) \citep{Nichol2021ImprovedModels} is the dominant method used for robotic manipulation. It uses a deterministic sampling process and outperforms  DDPM when using only a few (10-100) sampling iterations. DDIM can be formulated as a first-order ODE solver. In Diffusion Probabilistic Models-solver (DPM-solver) \citep{Lu2022DPM-Solver:Steps}, a second-order ODE solver is applied, which decreases the truncation error, thus further increasing performance on several image classification benchmarks for a low number of sampling steps. In contrast to DDIM, \cite{Karras2022ElucidatingModels,Lu2022DPM-Solver:Steps} use non-uniform step sizes in the solver. In a detailed analysis \cite{Karras2022ElucidatingModels} empirically shows that compared to uniform step-sizes, linear decreasing step sizes during denoising lead to increased performance \citep{Karras2022ElucidatingModels}, indicating that errors near the true distribution have a larger impact.

Even though DPM-solver \citep{Lu2022DPM-Solver:Steps} shows superior performance over DDIM. It should be noted that in the original papers \citep{Song2021DenoisingModels, Lu2022DPM-Solver:Steps}, only image-classification benchmarks are considered to compare both methods. Therefore, more extensive tests should be performed to validate these results.

A second group of methods addressing sampling speed also adapts the training process or requires additional fine-tuning. Examples are knowledge distillation of DMs to gradually reduce the number of noise levels \citep{Salimans2022PROGRESSIVEMODELS}, or finetuning of the noise schedule \citep{Nichol2021ImprovedModels, Watson2022LEARNINGQUALITY}. While in DDPM and DDIM, the noise schedule is fixed, in improved Denoising Diffusion Probabilistic Models (iDDPM) \citep{Nichol2021ImprovedModels}, the noise schedule is learned, resulting in better sample quality. 
They also suggest changing from a linear noise schedule, like in DDPM, to other schedules, e.g., a cosine noise schedule. In particular, for low-resolution samples, a linear schedule leads to a noisy diffusion process with too rapid information loss, while the cosine noise schedule has smaller steps during the beginning and end of the diffusion process. Already after a fraction of around 0.6 diffusion steps, the linear noise schedule is close to zero (and the data distribution close to white noise). Thus, the first steps of the reverse process do not strongly contribute to the data generation process, making the sampling process inefficient.
Although iDDPM \citep{Nichol2021ImprovedModels} also outperforms DDIM, it requires fine-tuning, which might be a reason why it is less popular.

There are also several methods \citep{Zhou2024VariationalExperts, Li2024CrosswayLearning, Wang2023ColdStates, Chen2024DontDiffusion} regarding sampling speed, specifically for applications in robotic manipulation, which is different from the previously named methodologies, which were developed in the context of image processing. For example, \cite{Chen2024DontDiffusion} samples from a more informed distribution than a Gaussian. They point out that even initial distributions approximated with simple heuristics result in better sample quality, especially when using few diffusion steps or when only a limited amount of data is available. Others \citep{Prasad2024ConsistencyDistillation} use teacher–student distillation techniques \citep{NIPS2017_68053af2}, where pretrained diffusion models serve as teachers, guiding student models to operate with larger denoising steps while preserving consistency with the teacher’s results at smaller steps. While this increases training effort, it decreases sampling time at inference, which is especially important in (near) real-time control.

Recently, flow matching \citep{lipman2022flow} has been used as an alternative method to diffusion. Like with diffusion, the true distribution is estimated starting from a noise distribution. However, instead of learning the time-dependent score or noise, and then deriving the velocity from noise to data distribution from it, in flow matching, the time-dependent velocity field is learned directly. This leads to a simpler training objective, using the interpolation between the noise sample and true data point, without requiring a noise schedule. Thus, flow matching is usually more numerically stable and requires less hyperparameter tuning. However, when using few sampling steps, with flow matching, there is a risk of mode-collapse and infeasible solutions, as the ODE-solver averages over the velocity field. Thus, \cite{frans2025one} conditions the model not only on the time-step, but also on the step-size. By using the fact that one large step should lead to the same point as two consecutive steps of half the size, they maximize a self-consistency objective in addition to the flow-matching objective. Thus, the model can sample with a single step, with only a small drop in performance, far surpassing the performance of DDIM, when only a small number of sampling steps are used. While this is similar to the above-mentioned distillation techniques \citep{Prasad2024ConsistencyDistillation}, here only a single model has to be trained.

\subsection{Adaptations for Robotic Manipulation}\label{chap:Adaptations}
Two main points must be considered to apply DMs to robotic manipulation. 
Firstly, in the diffusion processes described in the previous sections, given the initial noise, samples are generated solely based on the trained noise prediction network or conditional score network. However, robot actions are usually dependent on simulated or real-world observations with multi-modal sensory data and the robot's proprioception. Thus, the network used in the denoising process has to be conditioned on these observations \citep{Chi2023DiffusionDiffusion}. Encoding observations varies in different algorithms. Some use ground truth state information, such as object positions \citep{Ada2024DiffusionLearning}, and object features, like object sizes \citep{Mishra2023GenerativeModels, Mendez-Mendez2023EmbodiedPlanning}. In this case, sim-to-real transfer is challenging due to 
sensor inaccuracies, object occlusions, or other adversarial settings, e.g., lightning conditions, 
Therefore, most methods directly condition on visual observations, such as images \citep{Si2024Tilde:DeltaHand, Bharadhwaj2024TowardsPlans, Vosylius2024RenderCloning, Chi2023DiffusionDiffusion, Shi2023Waypoint-BasedManipulation}, point clouds \citep{Liu2023ComposableManipulation, Li2024GeneralizableGuidance}, or feature encodings and embeddings \citep{Ze20243DRepresentations, Ke20243DRepresentations, Li2024CrosswayLearning, Pearce2023ImitatingModels, Liang2023SkillDiffuser:Execution, Xian2023ChainedDiffuser:Manipulation, Xu2023XSkill:Discovery}, where the robustness to adversarial setting can be directly addressed.

\begin{figure}[t!]
    \centering
        \includegraphics[width=\textwidth]{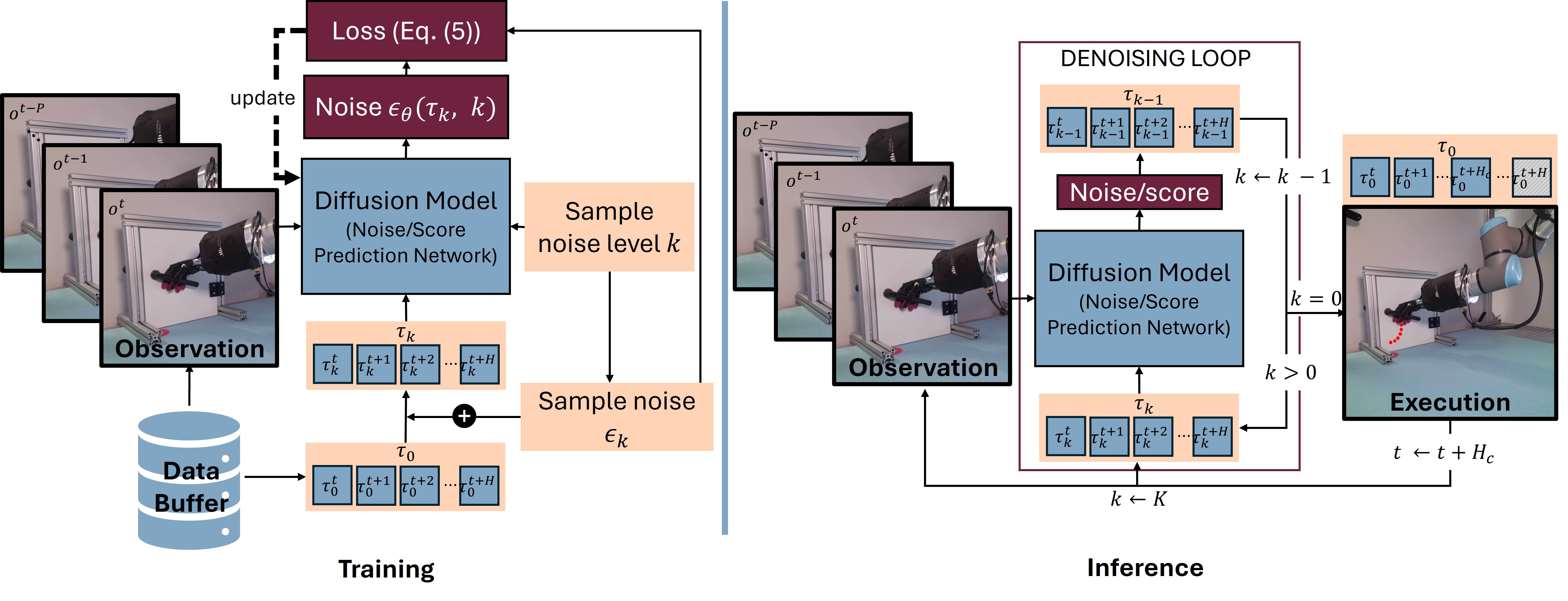}
        \caption{\small Illustrations of the iterative trajectory generation using receding horizon control. At inference, the trajectory is planned up to a planning horizon~$H$, conditioned on the past~$P$ observations~$\{o_t, o_{t-1}, \cdots, o_{t-P}\}$. Of this plan, only the steps until the control horizon~$H_c \leq H$ are executed. In the figure, this is visualized in the outer loop with the time variable~$t$. In the inner denoising loop, one subtrajectory~$\tau = \{\tau_t, \tau_{t + 1}, \cdots, \tau_{t + H}\}$ at the current time step~$t$ is generated, using a diffusion model. Conditioned on the last~$P$ observations and the current noise level~$k$, the diffusion model predicts the noise, or score, dependent on the model type. Using the predicted noise/score, the trajectory at the next lower noise level~$k - 1$ is calculated. This is then used as the next input to the diffusion model until the trajectory is completely denoised ($k = 0$), at which point it is executed. After execution of the subtrajectory, the time is increased and the next~$H$ steps of the trajectory are planned. For training, ground truth trajectories and corresponding observations are sampled from the data buffer. The diffusion model is also trained on subtrajectories. However, the lookahead~$H$ during training may be chosen larger than during inference, to ensure flexibility. The diffusion model is trained to predict the noise of a noisy trajectory. For this, first, a noise level~$k$ is sampled. Then the noise~$\epsilon_k$ is sampled, according to the predefined variance schedule. The noise is added in closed form to the ground-truth trajectory~$\tau_0$ (see \cref{eq:noising-closed-form}) to get the noisy trajectory~$\tau_k$. The predicted noise~$\epsilon_\theta(\tau_k, k)$ on the trajectory~$\tau_k$ is compared with the true sampled noise~$\epsilon_k$ to compute the loss. Using this, the diffusion model can be updated.}
        \label{fig:fig2}
\end{figure}

Secondly, unlike in image generation, where the pixels are spatially correlated, in trajectory generation for robotic manipulation, the samples of a trajectory are temporally correlated. On the one hand, generating complete trajectories may not only lead to high inaccuracies and error accumulation of the long-horizon predictions, but also prevent the model from reacting to changes in the environment. On the other hand, predicting the trajectory one action at a time increases the compounding error effect and may lead to frequent switches between modes. Accordingly,  trajectories are mostly predicted in subsequences, with a receding horizon, e.g., \citep{Chi2023DiffusionDiffusion, Scheikl2024MovementObjects}, which will be discussed in more detail in \cref{chap:traj-planning} and is visualized in \cref{fig:fig2}. In receding horizon control, the diffusion model generates only a subtrajectory with each backward pass. The subtrajectory is executed before generating the next subtrajectory on the updated observations. In comparison, grasps are generated similarly to images. As here only a single action, usually the grasp pose, is generated, this is done using a single backward pass of the diffusion model. Moreover, the grasp pose is usually predicted from a single initial observation. During execution, possible changes in the scene are not being taken into account. The backward pass for generating one action is visualized in \cref{fig:fig1}. 

\section{Architecture}\label{chap:architecture}
\subsection{Network Architecture}
For the implementation of the DM, it is essential to select an appropriate architecture for the noise prediction network.
There exist three predominant architectures used for the denoising diffusion networks: Convolutional neural networks (CNNs), transformers, and Multi-Layer Perceptrons (MLPs).

\subsubsection{Convolutional neural networks}
The most frequently employed architecture is the CNN, more specifically the Temporal U-Net that was first introduced by \cite{Janner2022PlanningSynthesis} in their algorithm Diffuser, a DM for robotics tasks. The U-Net architecture \citep{Ronneberger2015U-Net:Segmentation} has shown great success in image generation with DMs, e.g., \citep{Ho2020DenoisingModels, Dhariwal2021DiffusionSynthesis, Song2021Score-BasedEquations}. U-net, in general, is proven to be sample efficient and can even generalize well with small training datasets \citep{10.1007/978-3-031-15937-4_57, meyerhyperspectral}. Thus, it has been adapted to robotic manipulation by replacing two-dimensional spatial convolutions with one-dimensional temporal convolutions \citep{Janner2022PlanningSynthesis}.

The temporal U-Net is further adapted by \cite{Chi2023DiffusionDiffusion} in their CNN-based Diffusion Policy (DP) for robotic manipulation. While in Diffuser, the state and action trajectories are jointly denoised, only the action trajectories are generated in DP. To ensure temporal consistency, the diffusion process is conditioned on a history of observations using feature-wise linear modification (FiLM) \citep{Perez2018FiLM:Layer}.
This formulation allows for an extension to different and multiple conditions by concatenating them in feature space before applying FiLM \citep{Li2024CrosswayLearning, Si2024Tilde:DeltaHand, Ze20243DRepresentations, Li2024GeneralizableGuidance, Wang2024PoCo:Learning}. Moreover, it also enables the incorporation of constraints embedded with an MLP \citep{Ajay2023ISDecision-Making, Zhou2023AdaptiveModels, Power2023SamplingModels}.

Discussed in more detail in \cref{par:constraint-planning}, \cite{Janner2022PlanningSynthesis} formulates conditioning as inpainting, where during inferences at each denoising step, specific states from the currently being generated sample are replaced with states from the condition. For example, the final state of a generated trajectory may be replaced by the goal state, for goal-conditioning. This only affects the sampling process at inference and, thus, does not require any adaptations of the network architecture.
However, it only supports point-wise conditions, severely limiting its applications.
Multiple frameworks \citep{Saha2024EDMP:Planning, Carvalho2023MotionModels, Wang2023ColdStates, Ma2024HierarchicalManipulation} directly employ the temporal U-Net architecture introduced by \cite{Janner2022PlanningSynthesis}.
However, as this type of conditioning is highly limited in its applications, FiLM conditioning is more common. A different but less-used architecture incorporates conditions via cross-attention mapped to the intermediate layers of the U-Net \citep{Zhang2024LanguageTasks}, which is more complicated to integrate than FiLM conditioning. 

\subsubsection{Transformers}
Another commonly used architecture for the denoising network are transformers. A history of observations, the current denoising time step, and the (partially denoised) action are input tokens to the transformer. Additional conditions can be integrated via self-and cross-attention, e.g., \citep{Chi2023DiffusionDiffusion, Mishra2024ReorientDiff:Manipulation}. The exact architecture of the transformer varies across methods. The more commonly used model is a multi-head cross-attention transformer as the denoising network , e.g., \citep{Chi2023DiffusionDiffusion, Pearce2023ImitatingModels, Wang2023ColdStates, Mishra2024ReorientDiff:Manipulation}. Others \citep{Bharadhwaj2024Track2Act:Manipulation, Mishra2023GenerativeModels} use architectures based on the method Diffusion Transformers \citep{Peebles2023ScalableTransformers}, which is the first method combining DMs with transformer architectures. There are also less commonly used architectures, such as using the output tokens of the transformer as input to an MLP, which predicts the noise \citep{Ke20243DRepresentations}. 

For completeness, we provide a list of works, using transformer architectures: \citep{Chi2023DiffusionDiffusion, Pearce2023ImitatingModels, Scheikl2024MovementObjects, Wang2023ColdStates, Ze20243DRepresentations, Feng2024FFHFlow:Time, Bharadhwaj2024Track2Act:Manipulation, Mishra2023GenerativeModels, Liu2023StructDiffusion:Objects, Xu2024SetModels, Mishra2024ReorientDiff:Manipulation, Liu2023ComposableManipulation, Vosylius2024RenderCloning, Reuss2023Goal-ConditionedPolicies, Iioka2023MultimodalInstructions, Huang2025DiffusionDiffusion}.

\subsubsection{Multi-Layer Perceptrons}

Predominantly used for applications in RL, MLPs are employed as denoising networks, e.g., \citep{Suh2023FightingMatching, Ding2023ConsistencyLearning, Pearce2023ImitatingModels}, which take concatenated input features, such as observations, actions, and denoising time steps, to predict the noise.
Although the architectures vary, it is common to use a relatively small number of hidden layers (2-4) \cite{Wang2023DiffusionLearning, Kang2023EfficientLearning, Suh2023FightingMatching, Mendez-Mendez2023EmbodiedPlanning}, using e.g., Mish activation \citep{Misra2019Mish:Function}, following the first method \citep{Wang2023DiffusionLearning}, integrating DMs with Q-learning. It is important to note that most of these methods do not use visual input. An exception from this is \citep{Pearce2023ImitatingModels}, which also evaluates using high-resolution image inputs with an MLP-based DM. However, for this, a CNN-based image encoder is first applied to the raw image observation, before the encoding is fed to the DM.

\subsubsection{Comparison}
An ongoing debate exists concerning the relative merits of different architectural choices, with each architecture exhibiting distinct advantages and disadvantages. \cite{Chi2023DiffusionDiffusion} implemented both a U-Net-based and a transformer-based denoising network with the application of trajectory planning. They observed that the CNN-based model exhibits lower sensitivity to hyperparameters than transformers. Moreover, they report that when using positional control, the U-net results in a slightly higher success rate for some complex visual tasks, such as transport, tool hand, and push-t. On the other hand, U-nets may induce an over-smoothing effect, thereby resulting in diminished performance for high-frequency trajectories and consequently affecting velocity control. Thus, in these cases, transformers will likely lead to more precise predictions. Furthermore, transformer-based architectures have demonstrated proficiency in capturing long-range dependencies and exhibit notable robustness when handling high-dimensional data, surpassing the abilities of CNNs, which is particularly significant for tasks involving long horizons and high-level decision-making \citep{Janner2022PlanningSynthesis, Dosovitskiy2021ANSCALE}.

While MLPs typically exhibit inferior performance, especially when confronted with complex problems and high-dimensional input data, such as images, they often demonstrate superior computational efficiency, which facilitates higher-rate sampling and usually requires fewer computational resources. Due to their training stability, they are a commonly used architecture in RL. In contrast, U-Nets, and especially transformers, are characterized by substantial resource consumption and prolonged inference times, which may hinder their application in real-time robotics.\citep{Pearce2023ImitatingModels}

In summary, transformers are the most powerful architecture for handling high-dimensional input and output spaces, followed by CNNs, while MLPs have the highest computational efficiency. For processing visual data, such as raw images, an important task in robotic manipulation, a CNN or a Transformer architecture should be chosen. Also, while MLPs are most computationally efficient, real-time control is possible with the other two architectures, integrating, for example, receding horizon control \citep{Matteingley2011RecedingControl} in combination with a more efficient sampling process, like DDIM. 

\subsection{Number of sampling steps}
In addition to the network architecture, a crucial decision is the choice of the number of training and sampling iterations. As described in \cref{chap:arch-improv-and-adapt}, each sample must undergo iterative denoising over several steps, which can be notably time-consuming, especially in the context of employing larger denoising networks with longer inference durations, such as transformers. Within the framework of DDPM, the number of noise levels during training is equal to the number of denoising iterations at the time of inference. This hinders its use in many robotic manipulation scenarios, especially those necessitating real-time predictions. Consequently, numerous methodologies employ DDIM, where the number of sampling iterations during inference can be significantly reduced compared to the number of noise levels used during training. Common choices of noise levels are 50-100 during training, but only a subset of five to ten steps during inference\citep{Chi2023DiffusionDiffusion, Ma2024HierarchicalManipulation, Huang2025DiffusionDiffusion, Scheikl2024MovementObjects}. Only a few works used less sampling (3-4) \citep{Vosylius2024RenderCloning, Reuss2023Goal-ConditionedPolicies} or more (20-30) \citep{Mishra2024ReorientDiff:Manipulation, Wang2024PoCo:Learning} sampling steps. 
 \cite{Ko2024LearningCorrespondences} documented a slight decline in performance when the number of sampling steps is reduced to $10\%$ with DDIM \citep{Ko2024LearningCorrespondences}. Therefore, it is imperative to consider an appropriate trade-off between sample quality and inference time, tailored to the specific task requirements. Still, only a few evaluations exist that compare DDPM-based, DDIM-based, or other samplers for robotic manipulation, and further investigation is required.


\section{Applications}\label{chap:applications}
In this section, we explore the most dominant applications of DMs in robotic manipulation: trajectory generation for robotic manipulation, robotic grasping, and visual data augmentation for vision-based robotics manipulations.
\subsection{Trajectory generation}\label{chap:traj-planning}

%
%
Trajectory planning in robotic manipulation is vital for enabling robots to move from one point to another smoothly, safely, and efficiently while adhering to physical constraints, like speed and acceleration limits, as well as ensuring collision avoidance. Classical planning methods, like interpolation-based and sampling-based approaches, can have difficulty handling complex tasks or ensuring smooth paths. For instance, Rapidly Exploring Random Trees \citep{Martinez2023RapidlyEnvironments} might generate trajectories with sudden changes because of the discretization process. As already discussed in the introduction, although popular data-driven approaches, such as GMMs and EBMs, theoretically pertain to the ability to model multi-model data distributions, in reality, they show suboptimal behavior, such as biasing modes or lack of temporal consistency \citep{Chi2023DiffusionDiffusion}. 
In addition, GMMs can struggle with high-dimensional input spaces \citep{Ho2020DenoisingModels}. Increasing the number of components and covariances also increases the models' ability to model more complex distributions and capture complex and intricate movement patterns. However, this can negatively impact the smoothness of the generated trajectories, making GMMs highly sensitive to their hyperparameters. In contrast, denoising DMs have demonstrated exceptional performance in processing and generating high-dimensional data. Furthermore, the distributions generated by denoising DMs are inherently smooth \citep{Ho2020DenoisingModels, Sohl-Dickstein2015DeepThermodynamics, Chi2023DiffusionDiffusion}.
This makes DMs well-suited for complex, high-dimensional scenarios where flexibility and adaptability are required.
While most methodologies that apply probabilistic DMs to robotic manipulation focus on imitation learning, they have also been adapted to their application in RL, e.g., \citep{Janner2022PlanningSynthesis, Wang2023DiffusionLearning}. 

In the following sections, the methodologies of DMs for trajectory generation will be further discussed and categorized. We will first explain their applications in imitation learning, followed by a discussion on their use in reinforcement learning. For an overview of the method architectures in imitation learning, see \cref{tab:architecture-IL}, and for reinforcement learning, see \cref{tab:architecture-RL}.

\subsubsection{Imitation Learning}\label{chap:imitation-learning}

In imitation learning \citep{zare2024survey}, robots attempt to learn a specified task by observing multiple expert demonstrations. This paradigm, commonly known as Learning from Demonstrations (LfD), involves the robot observing expert examples and attempting to replicate the demonstrated behaviors. In this domain, the robot is expected to generalize beyond the specific demonstrations, which allows the robot to adapt to variations in tasks or changes in configuration spaces. This may include diverse observation perspectives, altered environmental conditions, or even new tasks that share structural similarities with those previously demonstrated. Thus, the robot must learn a representation of the task that allows flexibility and skill acquisition beyond the specific scenarios it was trained on.
Recent advancements in applying DMs to learn visuomotor policies \citep{Chi2023DiffusionDiffusion} enable the generation of smooth action trajectories by modeling the task as a generative process conditioned on sensory observations. Diffusion-based models, initially popularized for high-dimensional data generation such as images and natural languages, have demonstrated significant potential in robotics by effectively learning complex action distributions and generating multi-modal behaviors conditioned on task-specific inputs. For instance, combining with recent progress in multiview transformers \citep{gervet2023act3d, goyal2023rvt} that leverage the foundation model features \citep{radford2021learning, oquab2023dinov2}, 3D diffuser actor \citep{Ke20243DRepresentations} integrates multi-modal representations to generate the end-effector trajectories. As another example, GNFactor \citep{Ze2023GNFactor:Fields} renders multiview features from Stable Diffusion \citep{rombach2022high} to enhance 3d volumetric feature learning. Very similar to diffusion, recently \citep{10769838} flow-matching-based policies have emerged for trajectory generation, generally leading to a more stable training process with fewer hyperparameters, as already mentioned in \cref{chap:sampling-speed}. \cite{nguyen2025flowmplearningmotionfields} additionally includes second-order dynamics into the flow-matching objective, learning fields on acceleration and jerk to ensure smoothness of the generated trajectories.

In terms of the type of robotic embodiment, most works use parallel grippers or simpler end-effectors. However, few methods perform dexterous manipulation using DMs \citep{Si2024Tilde:DeltaHand, Ma2024DexDiff:Environments, Ze20243DRepresentations, Chen2024DontDiffusion, Wang2024DexCap:Manipulation, freiberg2024diffusion, welte2025interactiveimitationlearningdexterous}, to facilitate their stability and robustness, also in this high-dimensional setting.

In the following sections, we will first repeat the process of sampling actions for trajectory planning with DMs and discuss common pose representations.
Then we shortly address different visual data modalities, in particular 2D vs 3D visual observations.
Afterwards, we look at methods formulating trajectory planning as image generation, before looking at applications in hierarchical, multi-task, and constrained planning, also looking at multi-task planning with vision language action models (VLAs).  A visualization of the taxonomy is provided in \cref{tab:traj-plan}. More details on the individual method architectures are provided in \cref{tab:architecture-IL}.

\begin{table}[ht!]
\small
    \centering
    \renewcommand{\arraystretch}{1.3}
    \setlength{\tabcolsep}{6pt}
\begin{tabular}{|p{2.5cm}|>{\raggedright\arraybackslash}p{2.5cm}|p{4cm}|p{5.5cm}|}
\hline
\textbf{Perspective} & \textbf{Category} & \textbf{Subcategory} & \textbf{References} \\
\hline
\multirow{3}{2.5cm}{Methodological} 
& \multirow{3}{2.5cm}{Actions and pose representations}
&  Task Space §\ref{par:traj-pose-repr} 
& \cite{Chi2023DiffusionDiffusion, Pearce2023ImitatingModels, Ze20243DRepresentations, Ha2023ScalingAcquisition, Ke20243DRepresentations, Xu2023XSkill:Discovery, Li2024CrosswayLearning, Si2024Tilde:DeltaHand, Scheikl2024MovementObjects, Xian2023ChainedDiffuser:Manipulation, Liu2023ComposableManipulation} \\
\cline{3-4}
& & Joint Space §\ref{par:traj-pose-repr} & \cite{Carvalho2023MotionModels, Saha2024EDMP:Planning, Urain2023SE3-DiffusionFields:Diffusion, Ma2024HierarchicalManipulation}\\
\cline{3-4}
& & Image Space §\ref{par:traj-in-im-space} & \cite{Ko2024LearningCorrespondences, Yang2024LearningSimulators, Zhou2024RoboDreamer:Imagination, Vosylius2024RenderCloning, Du2023LearningGeneration, Liang2023SkillDiffuser:Execution}\\
\cline{2-4}
& \multirow{2}{2.5cm}{Visual data modality §\ref{par:vis-data-modality}}
& 2D & e.g. \cite{Chi2023DiffusionDiffusion, Liang2023SkillDiffuser:Execution, Scheikl2024MovementObjects, Si2024Tilde:DeltaHand}\\
\cline{3-4}
& & 3D & \cite{Li2024GeneralizableGuidance, Liu2023ComposableManipulation, Wang2024DexCap:Manipulation, Ze20243DRepresentations,Xian2023ChainedDiffuser:Manipulation, Ke20243DRepresentations}\\
\hline
\multirow{3}{2.5cm}{Functional} 
& \multirow{3}{2.5cm}{Long-Horizon and Multi-Task Learning} 
& Hierarchical Planning §\ref{par:traj-hier-planning}& \cite{Zhang2024LanguageTasks, Ma2024HierarchicalManipulation, Xian2023ChainedDiffuser:Manipulation, Ha2023ScalingAcquisition, Huang2024SubgoalManipulation, Du2023LearningGeneration} \\
\cline{3-4}
&  & Skill Learning §\ref{par:traj-hier-planning}
& \cite{Mishra2023GenerativeModels, Kim2024RobustDiffusion, Xu2023XSkill:Discovery, Liang2023SkillDiffuser:Execution}\\
\cline{3-4}
&& Vision Language Action Models §\ref{par:vlas} & \cite{pan2024visionlanguageactionmodeldiffusionpolicy, shentu2024llmsactionslatentcodes, octomodelteam2024octoopensourcegeneralistrobot,wen2024tinyvlafastdataefficientvisionlanguageaction, liu2024rdt1bdiffusionfoundationmodel,li2024cogactfoundationalvisionlanguageactionmodel,black2024pi0visionlanguageactionflowmodel}\\
\cline{2-4}
& \multirow{2}{2.5cm}{Constrained Planning §\ref{par:constraint-planning}}  
& Classifier guidance & \cite{Mishra2023GenerativeModels, Liang2023AdaptDiffuser:Planners, Janner2022PlanningSynthesis, Carvalho2023MotionModels} \\
\cline{3-4}
&& Classifier-free guidance & \cite{Ho2021Classifier-FreeGuidance, Saha2024EDMP:Planning, Li2024GeneralizableGuidance, Power2023SamplingModels, Reuss2024MultimodalGoals, Reuss2023Goal-ConditionedPolicies}\\
\hline
\end{tabular}
\caption{\small Taxonomy of Imitation Learning Approaches for Trajectory Generation with Diffusion Models}
    \label{tab:traj-plan}
\end{table}

\begin{table}[t]
\centering
\setlength{\tabcolsep}{4pt}
\small
\begin{tabular}{|l|p{3cm}|p{2cm}|p{3.5cm}|l|c|}
\hline
\textbf{Reference} & \textbf{Input} & \textbf{Output} & \textbf{Encoder} & \textbf{Diffuser} & \textbf{H} \\
\hline
\cite{Chi2023DiffusionDiffusion} & RGB$^{MV}$ & RHC & ResNet & FiLM & \bxmark \\
\arrayrulecolor{lightgray}\hline\arrayrulecolor{black}
\cite{Xian2023ChainedDiffuser:Manipulation} & RGB-D $^{MV}$, Lan & CT & CLIP & DiT \& MLP & \bcmark \\
\arrayrulecolor{lightgray}\hline\arrayrulecolor{black}
\cite{Reuss2023Goal-ConditionedPolicies} & GTS/RGB $^\text{SV}$ & CT & ResNet & DiT & \bxmark \\
\arrayrulecolor{lightgray}\hline\arrayrulecolor{black}
\cite{Chen2023PlayFusion:Play} & RGB $^\text{SV}$, Lan & RHC & ResNet & U-Net & \bxmark \\
\arrayrulecolor{lightgray}\hline\arrayrulecolor{black}
\cite{Zhou2023AdaptiveModels} & RGB $^{MV}$ & RHC & CLIP & U-Net & \bxmark \\
\arrayrulecolor{lightgray}\hline\arrayrulecolor{black}
\cite{Pearce2023ImitatingModels} & RGB $^\text{SV}$ & RHC & CNN/ResNet & MLP/DiT & \bxmark \\
\arrayrulecolor{lightgray}\hline\arrayrulecolor{black}
\cite{Mendez-Mendez2023EmbodiedPlanning} & GTS & RHC & - & MLPs & \bcmark  \\
\arrayrulecolor{lightgray}\hline\arrayrulecolor{black}
\cite{Ze20243DRepresentations} & PCs $^\text{SV}$ & RHC & MLP & FiLM & \bxmark \\
\arrayrulecolor{lightgray}\hline\arrayrulecolor{black}
\cite{Ke20243DRepresentations} & RGB-D$^\text{SV/MV}$, Lan & CT & CLIP & DiT & \bxmark \\
\arrayrulecolor{lightgray}\hline\arrayrulecolor{black}
\cite{Power2023SamplingModels} & GTS & RHC & MLP & U-Net & \bxmark \\
\arrayrulecolor{lightgray}\hline\arrayrulecolor{black}
\cite{Ma2024HierarchicalManipulation} & RGB-D $^\text{SV}$, Lan & J & PointNet++, MLP & U-Net & \bcmark \\
\arrayrulecolor{lightgray}\hline\arrayrulecolor{black}
\cite{Vosylius2024RenderCloning} & RGB $^{MV}$ & RHC & Transformer & DiT & \bxmark \\
\arrayrulecolor{lightgray}\hline\arrayrulecolor{black}
\cite{Zhang2024LanguageTasks} & RGB$^\text{SV}$, Lan & RHC & HULC, T5 & U-Net & \bcmark \\
\arrayrulecolor{lightgray}\hline\arrayrulecolor{black}
\cite{Reuss2024MultimodalGoals} & RGB $^{MV}$, Lan & RHC & ResNet, CLIP & DiT & \bxmark \\
\arrayrulecolor{lightgray}\hline\arrayrulecolor{black}
\cite{Scheikl2024MovementObjects} & RGB $^\text{SV}$/GTS & RHC & ResNet & DiT & \bxmark \\
\arrayrulecolor{lightgray}\hline\arrayrulecolor{black}
\cite{Chen2024DontDiffusion} & GTS/PCs/RGB$^\text{SV}$ & RHC & / & / & \bxmark \\
\arrayrulecolor{lightgray}\hline\arrayrulecolor{black}
\cite{Zhou2024VariationalExperts} & GTS/RGB$^\text{SV}$ & RHC & ResNet & DiT & \bcmark \\
\arrayrulecolor{lightgray}\hline\arrayrulecolor{black}
\cite{Li2024GeneralizableGuidance} & PCs$^\text{SV}$, Lan & RHC & SAM, XMem & FiLM & \bxmark \\
\arrayrulecolor{lightgray}\hline\arrayrulecolor{black}
\cite{Li2024CrosswayLearning} & RGB$^{MV}$ & RHC & ResNet & FiLM & \bxmark \\
\arrayrulecolor{lightgray}\hline\arrayrulecolor{black}
\cite{Si2024Tilde:DeltaHand} & RGB$^\text{SV}$ & RHC & ResNet & FiLM & \bxmark \\
\arrayrulecolor{lightgray}\hline\arrayrulecolor{black}
\cite{Saha2024EDMP:Planning} & GTS & RHC & - & U-Net & \bxmark \\
\arrayrulecolor{lightgray}\hline\arrayrulecolor{black}
\cite{Bharadhwaj2024Track2Act:Manipulation} & RGB$^\text{SV}$ & point tracks & / & DiT & \bxmark \\
\arrayrulecolor{lightgray}\hline\arrayrulecolor{black}
\cite{Wang2024PoCo:Learning} & RGB, Tactile, PCs, Lan & RHC & ResNet, PointNet, T5 & U-Net & \bxmark \\
\arrayrulecolor{lightgray}\hline\arrayrulecolor{black}
\cite{reuss2024efficient} & RGB, Lan & RHC & ResNet, CLIP & DiT & \bxmark \\
\hline
\end{tabular}

\caption{\small Technical details of trajectory diffusion using imitation learning. The references for the encoders are provided in \cref{tab:ref-encoder}. In the following, the symbols and abbreviations are explained:
H:~Whether the method is hierarchical (\bcmark) or not (\bxmark). PCs: Point Clouds, Lan: Language, GTS: Ground Truth State, and wether the visual input modality is from single view ($^\text{SV}$) or multi-view ($^\text{MV}$).
 U-Net:~temporal U-Net \citep{Janner2022PlanningSynthesis},
FiLM: Convolutional Neural Networks with Feature-wise Linear Modulation \cite{Perez2018FiLM:Layer}, DiT: Diffusion Transformer, RHC: sub-trajectories with receding horizon control, CT: complete trajectory in task space, J: complete trajectory in joint space.
A~``/'' indicates that the information is not provided by the cited paper, while a ``-'' indicates that no specialized encoder is required as ground truth state information is used.}
\label{tab:architecture-IL}
\end{table}

\paragraph{Actions and Pose Representation}\label{par:traj-pose-repr}

As briefly discussed in \cref{chap:Adaptations}, the entire trajectory can be generated as a single sample, multiple subsequences can be sampled using receding horizon control, or the trajectory can be generated by sampling individual steps. 
Only in a few methods \citep{Janner2022PlanningSynthesis, Ke20243DRepresentations} the whole trajectory is predicted at once. Although this enables a more efficient prediction, as the denoising has to be performed only once, it prohibits adapting to changes in the environment, requiring better foresight and making it unsuitable for more complex task settings with dynamic or open environments. 
On the other hand, sampling of individual steps increases the compounding error effect and can negatively affect temporal correlation.
Instead of predicting micro-actions, some use DMs to predict waypoints \citep{Shi2023Waypoint-BasedManipulation}. This can decrease the compounding error, by reducing the temporal horizon. However, it relies on preprocessing or task settings that ensure that the space in between waypoints is not occluded.
Thus, typically, DMs generate trajectories consisting of sequences of micro-actions represented as end-effector positions, generally encompassing translation and rotation depending on end-effector actuation \citep{Chi2023DiffusionDiffusion, Ze20243DRepresentations, Xu2023XSkill:Discovery, Li2024CrosswayLearning, Si2024Tilde:DeltaHand, Scheikl2024MovementObjects, Ke20243DRepresentations, Ha2023ScalingAcquisition}. Once the trajectory is sampled, the proximity of the predicted positions enables computing the motion between the positions with simple positional controllers without the need for complex trajectory planning techniques. The control scheme is visualized in detail in \cref{fig:fig2}. Although more commonly applied in grasp prediction, here the pose is sometimes also represented in special Euclidean group ($\mathbf{SE}(3)$) \citep{Xian2023ChainedDiffuser:Manipulation, Liu2023ComposableManipulation, Ryu2023Diffusion-EDFs:Manipulation}. Explained in more detail in \cref{chap:grasp-learning}, the group structure of the $\mathbf{SE}(3)$ Lie group enables continuous interpolation and transformations between multiple object poses. As \citep{Liu2023ComposableManipulation, Ryu2023Diffusion-EDFs:Manipulation} performs complex tasks involving trajectory planning and grasping for aligning multiple objects, these properties are important to ensure physically and geometrically grounded actions. However, as the prediction of~$\mathbf{SE}(3)$ poses with DMs requires a more complex model structure and training in imitation learning, it is more usual to use representations, such as Euler angles or quaternions, in trajectory planning. Not only diffusion, but also flow matching has been adapted to use representations in~~$\mathbf{SE}(3)$ or Riemannian manifolds in general \citep{10801521}.

Although not common,  sometimes actions are
predicted directly in joint space \citep{Carvalho2023MotionModels, Pearce2023ImitatingModels, Saha2024EDMP:Planning, Ma2024HierarchicalManipulation}, allowing for direct control of joint motions, which, e.g., reduces singularities.




\paragraph{Visual Data Modalities}\label{par:vis-data-modality}
As already discussed in \cref{chap:Adaptations} to ground the robots actions in the physical world, they are dependent on sensory input. Here, in the majority of methods visual observations are used.
In the original work \citep{Chi2023DiffusionDiffusion}, combining visual robotic manipulation with DMs for trajectory planning, the DM is conditioned on RGB-image observations. Many methods, e.g., \citep{Si2024Tilde:DeltaHand, Pearce2023ImitatingModels, Li2024CrosswayLearning}, adopt using RGB inputs, also developing more intricate encoding schemes \citep{qi2025ecdiffusermultiobjectmanipulationentitycentric}.

However, 2D visual scene representations may not provide sufficient geometrical information for intricate robotic tasks, especially in scenes containing occlusions.
Thus, multiple later methods used 3D scene representations instead. Here, DMs are either directly conditioned on the point cloud \citep{Li2024GeneralizableGuidance, Liu2023ComposableManipulation, Wang2024DexCap:Manipulation} or point cloud feature embeddings \citep{Ze20243DRepresentations,Xian2023ChainedDiffuser:Manipulation, Ke20243DRepresentations}, from singleview \citep{Ze20243DRepresentations, Li2024GeneralizableGuidance, Wang2024DexCap:Manipulation}, or multiview camera setups \citep{Ke20243DRepresentations, Xian2023ChainedDiffuser:Manipulation}.
While multiview camera setups provide more complete scene information, they also require a more involved setup and more hardware resources.

These models outperform methods relying solely on 2D visual information, on more complex tasks, also demonstrating robustness to adversarial lighting conditions. 

\paragraph{Trajectory Planning as Image Generation}\label{par:traj-in-im-space}
Another category formulates trajectory generation directly in image space, leveraging the exceptional generative abilities of DMs in image generation. Here  \citep{Ko2024LearningCorrespondences, Zhou2024RoboDreamer:Imagination, Du2023LearningGeneration}, given a single image observation, a sequence of images, or a video, sometimes in combination with a language-task-instruction, the diffusion process is conditioned to predict a sequence of images, depicting the change in robot and object position.
This comes with the benefit of internet-wide video training data, which facilitates extensive training, leading to good generalization capabilities. Especially in combination with methods \citep{Bharadhwaj2024Track2Act:Manipulation} agnostic to the robot embodiment, this highly increases the amount of available training data. Moreover, in robotic manipulation, the model usually has to parse visual observations. Predicting actions in image space circumvents the need for mapping from the image space to a usually much lower-dimensional action space, reducing the required amount of training data \citep{Vosylius2024RenderCloning}. However, predicting high-dimensional images may also prevent the model from successfully learning important details of trajectories, as the DM is not guided to pay more attention to certain regions of the image, even though usually only a low fraction of pixels contain task-relevant information. Additionally, methods generating complete images must ensure temporal consistency and physical plausibility. Hence, extensive training resources are required. As an example, \citep{Zhou2024RoboDreamer:Imagination} uses 100 V100 GPUs and 70k demonstrations for training.
While still operating in image space, some methods do not generate whole image sequences, but instead perform point-tracking \citep{Bharadhwaj2024Track2Act:Manipulation} or diffuse imprecise action-effects on the end-effector position directly in image space \citep{Vosylius2024RenderCloning}. This mitigates the problem of generating physically implausible scenes. However, point-tracking still requires extensive amounts of data. \cite{Bharadhwaj2024Track2Act:Manipulation}, e.g., uses 0.4 million video clips for training.

\paragraph{Long-Horizon and Multi-Task Learning}\label{par:traj-hier-planning}
Due to their ability to robustly model multi-model distributions and relatively good generalization capabilities, DMs are well suited to handle long-horizon and multi-skill tasks, where usually long-range dependencies and multiple valid solutions exist, especially for high-level task instructions \citep{Mendez-Mendez2023EmbodiedPlanning, Liang2023SkillDiffuser:Execution}. 
Often, long-horizon tasks are modeled using hierarchical structures and skill learning. Usually, a single skill-conditioned DM or several DMs are learned for the individual skills, while the higher-level skill planning does not use a DM \citep{Mishra2023GenerativeModels, Kim2024RobustDiffusion, Xu2023XSkill:Discovery, Liang2023SkillDiffuser:Execution}. The exact architecture for the higher-level skill planning varies across methods, being, for example, a variational autoencoder \citep{Kim2024RobustDiffusion} or a regression model \citep{Mishra2023GenerativeModels}.
Instead of having a separate skill planner that samples one skill, \cite{Wang2024PoCo:Learning} develops a sampling scheme that can sample from a combination of DMs trained for different tasks and in different settings.

To forego the skill-enumeration, which brings with it the limitation of a predefined finite number of skills, some works employ a coarse-to-fine hierarchical framework, where higher-level policies are used to predict goal states for lower-level policies \citep{Zhang2024LanguageTasks, Ma2024HierarchicalManipulation, Xian2023ChainedDiffuser:Manipulation, Ha2023ScalingAcquisition, Huang2024SubgoalManipulation, Du2023LearningGeneration}.

The ability of DMs to stably process high-dimensional input spaces enables the integration of multi-modal inputs, which is especially important in multi-skill tasks, to develop versatile and generalizable agents via arbitrary skill-chaining. Methodologies use videos \citep{Xu2023XSkill:Discovery}, images, and natural language task instructions \citep{Liang2023SkillDiffuser:Execution, Wang2024PoCo:Learning, Zhou2024RoboDreamer:Imagination, reuss2024efficient}, or even more diverse modalities, such as tactile information and point clouds \citep{Wang2024PoCo:Learning}, to prompt skills.

Although these methods are designed to enhance generalizability, achieving adaptability in highly dynamic environments and unfamiliar scenarios may require the integration of continuous and lifelong learning. This is a widely unexplored field in the context of DMs, with only very few works \citep{huang2024embodiedgeneralistagent3d, DiPalo2024DiffusionLearning} exploring this topic. Moreover, these methods are still limited in their applications. \citep{DiPalo2024DiffusionLearning} are utilizing a lifelong buffer to accelerate the training of new policies for new tasks. In contrast, \citep{Mendez-Mendez2023EmbodiedPlanning} continually updates its policy. However, they only conduct training and experiments in simulation. Additionally, their method requires precise feature descriptions of all involved objects and is limited to predefined abstract skills. Moreover, for the continual update, all past data is replayed, which is not only computationally inefficient but also does not prevent catastrophic forgetting.

\paragraph{Multi-Task Learning with Vision Language Action Models}\label{par:vlas}

Another approach to enhance generalizability in multi-task settings is the incorporation of pretrained VLAs. As a specialized class of multimodal language model (MLLM), VLAs combine the perceptual and semantic representation power of the vision language foundation model and the motor execution capabilities of the action generation
model, thereby forming a cohesive end-to-end decision-making framework. Being pretrained on internet-scale data, VLAs exhibit great generalization capabilities across diverse and unseen scenarios, thereby enabling robots to execute
complex tasks with remarkable adaptability \citep{firoozi2023foundationmodelsroboticsapplications}.

A predominant line of approaches among VLAs employs next-token prediction for auto-regressive action token generation, representing a foundational approach to end-to-end VLA modeling, e.g.,  \citep{brohan2023rt1roboticstransformerrealworld,brohan2023rt2visionlanguageactionmodelstransfer,kim2024openvlaopensourcevisionlanguageactionmodel}. However, this approach is hindered by significant limitations, most notably the slow inference speeds inherent to auto-regressive methods \citep{brohan2023rt2visionlanguageactionmodelstransfer,wen2024tinyvlafastdataefficientvisionlanguageaction,pertsch2025fastefficientactiontokenization}. This poses a critical bottleneck for real-time robotic systems, where low-latency decision-making is essential. Furthermore, the discretizations of motion tokens, which reformulates action generation as a classification task, introduces quantization errors that lead to a decrease in control precision, thus reducing the overall performance and reliability \citep{zhang2024grapegeneralizingrobotpolicy,Pearce2023ImitatingModels,zhang2024vlabenchlargescalebenchmarklanguageconditioned}.

To address these limitations one line of research within VLAs focuses on predicting future states and synthesizing executable actions by leveraging inverse kinematics principles derived from these predictions, e.g., \citep{cheang2024gr2generativevideolanguageactionmodel,zhen20243dvla3dvisionlanguageactiongenerative,zhang2024navidvideobasedvlmplans}. While this approach addresses some of the limitations associated with token discretization, multimodal states often correspond to multiple valid actions, and the attempt to model these states through techniques such as arithmetic averaging can result in infeasible or suboptimal action outputs.

Thus, showing strong capabilities and stability in modeling multi-modal distributions, DMs have emerged as a promising solution.
Leveraging their strong generalization capabilities, a VLA is used to predict coarse action, while a DM-based policy refines the action, to increase precision and adaptability to different robot embodiments, e.g. \citep{pan2024visionlanguageactionmodeldiffusionpolicy, shentu2024llmsactionslatentcodes, octomodelteam2024octoopensourcegeneralistrobot}.
For instance, TinyVLA \citep{wen2024tinyvlafastdataefficientvisionlanguageaction} incorporates a diffusion-based head module on top of a pretrained VLA to directly generate robotic actions. 
More specifically, DP \citep{Chi2023DiffusionDiffusion} is connected to the multimodal model backbone via two linear projections and a LayerNorm.
The multimodal model backbone jointly encodes the current observations and language instruction, generating a multimodal embedding that conditions and guides the denoising process. Furthermore, in order to better fill the gap between logical reasoning and actionable robot policies, a reasoning injection module is proposed, which reuses reasoning outputs\citep{wen2024diffusionvlascalingrobotfoundation}. Similarly, conditional diffusion decoders have been leveraged to represent continuous multimodal action distributions, enabling the generation of diverse and contextually appropriate action sequences \citep{octomodelteam2024octoopensourcegeneralistrobot, liu2024rdt1bdiffusionfoundationmodel,li2024cogactfoundationalvisionlanguageactionmodel}. 

Addressing the disadvantage of long inference times with DMs, in some recent works instead, flow matching is used to generate actions from observations preprocessed by VLMs to solve flexible and dynamic tasks, offering a robust alternative to traditional diffusion mechanisms \citep{black2024pi0visionlanguageactionflowmodel, zhang2025affordancebasedrobotmanipulationflow}. While \cite{black2024pi0visionlanguageactionflowmodel} takes a skill-based approach, where the vision-language model is used to decide on actions, \cite{zhang2025affordancebasedrobotmanipulationflow} uses a vision-language model to generate waypoints. In both approaches, flow matching is used as the expert policy, generating precise trajectories. 

VLAs offer access to models trained on huge amounts of data and with strong computational power, leading to strong generalization capabilities. To mitigate some of their shortcomings, such as imprecise actions, specialized policies can be used for refinement. To not restrict the generalizability of the VLA, DMs offer a great possibility, as they can capture complex multi-model distributions and process high-dimensional visual inputs. However, both VLAs and DMs have a relatively slow inference speed. Thus, especially in this combination with VLAs, increasing the sampling efficiency of DMs is important. One example was provided in the previous paragraph. But the topic of higher sampling speed with DMs is also discussed in more detail in \cref{chap:sampling-speed}.


\paragraph{Constrained planning}\label{par:constraint-planning}
Another line of methods focuses on constrained trajectory learning. A typical goal is obstacle avoidance, object-centric, or goal-oriented trajectory planning, but other constraints can also be included.
If the constraints are known prior to training, they can be integrated into the loss function. However, if the goal is to adhere to various and possibly changing constraints during inference, another approach has to be taken.
For less complex constraints, such as specific initial or goal states, \citep{Janner2022PlanningSynthesis} introduces a conditioning, where, after each denoising time step (\cref{eq:denoising-process}), the particular state from the trajectory is replaced by the state from the constraint. However, this can lead the trajectory into regions of low likelihood, hence decreasing stability and potentially causing mode collapse. Moreover, this method is not applicable to more complex constraints. 
 
 One approach, also addressed by \cite{Janner2022PlanningSynthesis}, is classifier guidance \citep{Dhariwal2021DiffusionSynthesis}. Here, a separate model is trained to score the trajectory at each denoising step and steer it toward regions that satisfy the constraint. This is integrated into the denoising process by adding the gradient of the predicted score. It should be noted that for sequential data, such as trajectories, classifier guidance can also bias the sampling towards regions of low likelihood \citep{Pearce2023ImitatingModels}. Thus, the weight of the guidance factor must be carefully chosen. Moreover, during the start of the denoising process the guidance model must predict the score on a highly uninformative output (close to Gaussian noise) and should have a lower impact. 
 Therefore, it is important to inform the classifier of the denoising time step, train it also on noisy samples, or adjust the weight with which the guidance factor is integrated into the reverse process.
Classifier guidance is applied in several methodologies \citep{Mishra2023GenerativeModels, Liang2023AdaptDiffuser:Planners, Janner2022PlanningSynthesis, Carvalho2023MotionModels}.
However, it requires the additional training of a separate model. Furthermore, computing the gradient of the classifier at each sampling step adds additional computational cost.
Thus, classifier-free guidance \citep{Ho2021Classifier-FreeGuidance, Saha2024EDMP:Planning, Li2024GeneralizableGuidance, Power2023SamplingModels, Reuss2024MultimodalGoals, Reuss2023Goal-ConditionedPolicies} has been introduced, where a conditional and an unconditional DM per constraint are trained in parallel. During sampling, a weighted mixture of both DMs is used, allowing for arbitrary combinations of constraints, also not seen together during training. However, it does not generalize to entirely new constraints, as this would necessitate the training of new conditional DMs.

As both classifier and classifier-free guidance only steer the training process, they do not guarantee constraint satisfaction.
To guarantee constraint satisfaction in delicate environments, such as surgery \citep{Scheikl2024MovementObjects}, incorporate movement primitives with DMs to ensure the quality of the trajectory.
Recent advances in diffusion models also delve into constraint satisfaction \citep{Romer2024DiffusionConstraints}, integrating constraint tightening into the reverse diffusion process. While this outperforms previous methods \citep{Power2023SamplingModels, Janner2022PlanningSynthesis, carvalho2024grasp} in regards to constraint satisfaction, also in multi-constraint settings and constraints not seen during training, the evaluation is done only in simulation on a single experiment setup. Thus, constraint satisfaction with DMs remains an interesting research direction to further explore.

Few methods also perform affordance-based optimization for trajectory planning \citep{Liu2023ComposableManipulation}. However, most work in affordance-based manipulation concentrates on grasp learning, which is discussed in more detail in \cref{chap:grasp-learning}.

\subsubsection{Offline Reinforcement Learning}\label{chap:RL}
To apply diffusion policies in the context of RL the reward term has to be integrated. Diffuser \citep{Janner2022PlanningSynthesis}, one early work adapting diffusion to RL, uses classifier-based guidance, which is based on classifier guidance described in \cref{par:constraint-planning}. Let~$\tau = \{(s_0, a_0), \dots, (s_T, a_T)\}$ be a trajectory with one state-action pair per timestep in a planning horizon~$\{0, \dots, T\}$. To incorporate the reward term during sampling, a regression model~$R_\phi(\tau_k)$ is trained to predict the return, i.e., the cumulative future reward, over the trajectory~$\tau_k$ at each denoising time step~$k \in \{0, \dots, K\}$. This is incorporated into the sampling process by adding the guidance term at each iteration of the reverse diffusion process \citep{Janner2022PlanningSynthesis}:

\begin{equation}
    p(\tau_{k-1} \mid \tau_k, \mathcal{O}_{1:T}) \approx \mathcal{N}(\tau_{k - 1}; \mu + \Sigma \nabla R_\phi(\mu), \Sigma ).
\end{equation}
Moreover, to ensure that the current state observation~$s_0$ is not changed by the reverse diffusion on the trajectory, $\tau_{s_0}^{k-1}$ is set to the current state observation after each reverse diffusion iteration. In the same way, goal-conditioning or other constraints, which can be accomplished by replacing states from the trajectory with states from the constraint, can be integrated into the method. This, is done in several methodologies \citep{Janner2022PlanningSynthesis,Liang2023AdaptDiffuser:Planners}. However, it has to be done with care, as it can lead to trajectories in regions of low likelihood which may cause instability and mode-collapse \citep{Janner2022PlanningSynthesis, Song2021Score-BasedEquations}.
After the reverse process is completed and~$\tau^0$ has been predicted, the first action~$a_0$ of the plan is executed. Then, the planning horizon is shifted one step forward, and the next action is sampled. 

In Diffuser \citep{Janner2022PlanningSynthesis} and Diffuser-based methods \citep{Suh2023FightingMatching, Liang2023AdaptDiffuser:Planners}, the DM is trained independently of the reward signal, similar to methods in imitation learning with DM. Not leveraging the reward signal for training the policy can lead to misalignment of the learned trajectories with optimal trajectories and thus suboptimal behavior of the policy. In contrast,  leveraging the reward signal already during training of the policy, can steer the training process, consequently increasing both quality of the trained policy and sample efficiency.

To mitigate these shortcomings, one approach, Decision Diffuser \citep{Ajay2023ISDecision-Making}, directly conditions the DM on the return of the trajectory using classifier-free guidance. This method outperforms Diffuser on a variety of tasks, such a block-stacking task. However, both methods have not been evaluated on real-world tasks.
Directly conditioning on the return, limits generalization capabilities. Different to Q-learning, where the value function is approximated, which generalizes across all future trajectories, here only the return of the current trajectory is considered. Sharing some similarity to on-policy methods, this limits generalization as the policy learns to follow trajectories from the demonstrations with high return values. Thus, this can also be interpreted as guided imitation learning.  

A more common method \citep{Wang2023DiffusionLearning} integrates offline Q-learning with DMs.
The loss function from \cref{eq:DDPM-loss} is a behavior cloning loss, as the goal is to minimize error with respect to samples taken via the behavior policy. \cite{Wang2023DiffusionLearning} suggests including a critic in the training procedure, which they call Diffusion Q-learning (Diffusion-QL). In Diffusion-QL a Q-function is trained, by minimizing the Bellman-Operator using the double Q-learning trick. The actions for updating the Q-function are sampled from the DM. In turn a policy improvement step $\mathcal{L}_c = - \mathbb{E}_{\boldsymbol{s}\sim\mathcal{D},\boldsymbol{a}^0\sim\pi_\theta}\left[Q_\phi(\boldsymbol{s},\boldsymbol{a}^0)\right]$ is included in the loss for updating the DM \citep{Wang2023DiffusionLearning}:
\begin{equation}
\begin{split}
\pi=\arg\min_{\pi_\theta}\mathcal{L}_{RL} = \arg\min_{\pi_\theta}\mathcal{L} + \alpha\mathcal{L}_c,
\end{split}
\end{equation}
where~$\mathcal{L}$ is the diffusion loss from \cref{eq:DDPM-loss} and the parameter~$\alpha$ regulates the influence of the critic.
Several methods \citep{Ada2024DiffusionLearning, Kim2024StitchingRL, Venkatraman2023ReasoningLearning, Kang2023EfficientLearning}, build on Diffusion Q-learning. To increase the generalizability to out-of-distribution data, a common problem in offline RL \citep{DBLP:journals/corr/abs-2005-01643}, \cite{Ada2024DiffusionLearning}, include a state-reconstruction loss, into the training of the DM. An overview of the architectures of methods combining diffusion and reinforcement learning is provided in \cref{tab:architecture-RL}.

\begin{table}[t]
\centering
\setlength{\tabcolsep}{4pt}
\small
\begin{tabular}{|l|p{2.2cm}|p{2.2cm}|p{3cm}|p{3cm}|c|}
\hline
\textbf{Reference} & \textbf{Input} & \textbf{Output} & \textbf{Encoder} & \textbf{Diffuser} & \textbf{H/S} \\
\hline
\cite{Janner2022PlanningSynthesis} & GTS & RHC & - & U-Net & \bxmark \\
\arrayrulecolor{lightgray}\hline\arrayrulecolor{black}
\cite{Ajay2023ISDecision-Making} & GTS & RHC & - & U-Net & \bcmark \\
\arrayrulecolor{lightgray}\hline\arrayrulecolor{black}
\cite{Wang2023DiffusionLearning} & GTS & SiA & - & MLP & \bxmark \\
\arrayrulecolor{lightgray}\hline\arrayrulecolor{black}
\cite{Wang2023ColdStates} & GTS & RHC & - & DiT & \bxmark \\
\arrayrulecolor{lightgray}\hline\arrayrulecolor{black}
\cite{Ding2023ConsistencyLearning} & GTS & SiA & - & MLP & \bxmark \\
\arrayrulecolor{lightgray}\hline\arrayrulecolor{black}
\cite{Mishra2023GenerativeModels} & GTS & RHC & - & DiT & \bcmark \\
\arrayrulecolor{lightgray}\hline\arrayrulecolor{black}
\cite{Kang2023EfficientLearning} & GTS & RHC & - & MLP & \bxmark \\
\arrayrulecolor{lightgray}\hline\arrayrulecolor{black}
\cite{Brehmer2023EDGI:Agents} & GTS & RHC & - & Eq. U-Net & \bxmark \\
\arrayrulecolor{lightgray}\hline\arrayrulecolor{black}
\cite{Suh2023FightingMatching} & GTS & RHC & - & U-Net & \bxmark \\
\arrayrulecolor{lightgray}\hline\arrayrulecolor{black}
\cite{Ha2023ScalingAcquisition} & RGB$^{MV}$, Lan & RHC & ResNet, CLIP & FiLM & \bcmark \\
\arrayrulecolor{lightgray}\hline\arrayrulecolor{black}
\cite{Kim2024StitchingRL} & GTS & RHC & - & U-Net & \bcmark \\
\arrayrulecolor{lightgray}\hline\arrayrulecolor{black}
\cite{Liang2023AdaptDiffuser:Planners} & GTS & RHC & - & U-Net & \bxmark \\
\arrayrulecolor{lightgray}\hline\arrayrulecolor{black}
\cite{Ada2024DiffusionLearning} & GTS & SiA & - & MLP & \bxmark \\
\arrayrulecolor{lightgray}\hline\arrayrulecolor{black}
\cite{Ren2024DiffusionOptimization} & RGB/GTS & SiA & ViT/- & U-Net/MLP & \bxmark \\
\arrayrulecolor{lightgray}\hline\arrayrulecolor{black}
\cite{Huang2025DiffusionDiffusion} & RGB$^{SV}$ & SiA & VQ-GAN & VQ-Diffusion \cite{9879180} & \bxmark \\
\arrayrulecolor{lightgray}\hline\arrayrulecolor{black}
\cite{Carvalho2023MotionModels} & GTS & RHC & - & U-Net & \bxmark \\
\hline
\end{tabular}

\caption{\small Technical details of trajectory diffusion using reinforcement learning. The references for the encoders are provided in \cref{tab:ref-encoder}. In the following, the symbols and abbreviations are explained:
H/S:~Whether the method is hierarchical/skill-based (\bcmark) or not (\bxmark). Lan: Language, GTS: Ground Truth State, and wether the visual input modality is from single view ($^\text{SV}$) or multi-view ($^\text{MV}$).
 U-Net:~temporal U-Net \citep{Janner2022PlanningSynthesis},
 Eq.:~Equivariant
FiLM: Convolutional Neural Networks with Feature-wise Linear Modulation \cite{Perez2018FiLM:Layer}, DiT: Diffusion Transformer, RHC: sub-trajectories with receding horizon control, Sia~=single actions.
A ``-'' indicates that no specialized encoder is required as ground truth state information is used.
}
\label{tab:architecture-RL}
\end{table}

One characteristic of methodologies combining RL with DMs is that they are offline methods, with both the policy, i.e., the DM, and the return prediction model/critic being trained offline. This introduces the usual advantages and disadvantages of offline RL \citep{DBLP:journals/corr/abs-2005-01643}. The model relies on high-quality existing data, consisting of state-action-reward transitions, and is unable to react to distribution shifts. If not tuned well, this may also lead to overfitting. On the other hand, it has increased sample efficiency and does not require real-time data collections and training, which decreases computational cost and can increase training stability. Compared to imitation learning \citep{DBLP:journals/corr/abs-2005-01643, 10886242, NIPS2016_cc7e2b87}, offline RL requires data labeled with rewards, the training of a reward function, and is more prone to overfitting to suboptimal behavior. However, confronted with data containing diverse and suboptimal behavior, offline RL has the potential of better generalization compared to imitation learning, as it is well suited to model the entire state-action space. Thus, combining RL with DMs has the potential of modeling highly multi-modal distributions over the whole state-action space, strongly increasing generalizability \citep{Liang2023AdaptDiffuser:Planners, Ren2024DiffusionOptimization}. In contrast, if high-quality expert demonstrations are available, imitation learning might lead to better performance and computational efficiency. 
To overcome some of the shortcoming of imitation learning, such as the covariate shift problem \citep{pmlr-v9-ross10a}, which make it difficult to handle out of distribution situations, some strategies are devised to finetune behavior cloning policies using RL \citep{Ren2024DiffusionOptimization, Huang2025DiffusionDiffusion}.

Skill-composition is a common method, to handle long-horizon tasks. To leverage the abilities of RL to learn from suboptimal behaviors multiple methodologies \citep{Ajay2023ISDecision-Making, Kim2024RobustDiffusion, Venkatraman2023ReasoningLearning, Kim2024StitchingRL} combine skill-learning and RL with DMs.

Only little research \citep{Ding2023ConsistencyLearning, Ajay2023ISDecision-Making} in online and offline-to-online RL with DMs has been conducted, leaving a wide field open for research. Moreover, in the context of skill-learning \citep{Ajay2023ISDecision-Making}, the DMs, used for the lower-level policies, are trained offline and remain frozen, while the higher-level policy are trained using online RL.

It should be noted that, apart from \cite{Ren2024DiffusionOptimization, Huang2025DiffusionDiffusion}, none of the aforementioned methods process visual observations and instead rely on ground-truth environment information, which is only easily available in simulation.
Moreover, while all methods have also been tested on robotic manipulation tasks, only a few \citep{Ren2024DiffusionOptimization, Huang2025DiffusionDiffusion} have been deliberately engineered for these specific applications. Expanding the scope to encompass all methodologies devised for robotics at large, there is a more substantial body of work that integrates diffusion policies with RL.


\subsection{Robotic grasp generation}\label{chap:grasp-learning}

\begin{table}[t]
    \small
    \centering
    \renewcommand{\arraystretch}{1.3}
    \setlength{\tabcolsep}{6pt}
\begin{tabular}{|p{2.5cm}|>{\raggedright\arraybackslash}p{2.5cm}|p{4cm}|p{5.5cm}|}
\hline
\textbf{Perspective} & \textbf{Category} & \textbf{Subcategory} & \textbf{References} \\
\hline
\multirow{3}{2.5cm}{Methodological} 
& \multirow{3}{2.5cm}{Diffusion on \(\mathbf{SE}(3)\) grasp poses}
& Parallel jaw grasp & \cite{Urain2023SE3-DiffusionFields:Diffusion, song2024implicit, singh2024constrained, lim2024equigraspflow, carvalho2024grasp,Ryu2023Diffusion-EDFs:Manipulation, freiberg2024diffusion,huang2025hgdiffuser}\\
\cline{3-4}
& & Dextrous grasp & \cite{Wu2024LearningGrasping, Weng2024DexDiffuser:Models, Wang2024Single-ViewGeneration, freiberg2024diffusion, zhong2025gagrasp,zhang2024dexgrasp, wu2023learning}\\
\cline{2-4}
& Diffusion in latent space & - &\cite{Barad2023GraspLDM:Models}\\
\cline{2-4}
& Diffusion as feature encoders \& image generators  & - & \cite{Li2024ALDM-Grasping:Grasping, Tsagkas2024ClickDescriptors}\\
\hline
\multirow{3}{2.5cm}{Functional} 
& \multirow{2}{2.5cm}{Affordance-driven diffusion} 
& Language-guided grasp diffusion & \cite{Nguyen2024LightweightModel, Vuong2024Language-drivenDetection, Nguyen2024Language-ConditionedClouds, Chang2024Text2Grasp:Parts, Zhang2024NL2Contact:Model}\\
\cline{3-4}
& & Pre-grasp manipulation via imitation learning & \cite{Wu2024Unidexfpm:Policy, Ma2024DexDiff:Environments}\\
\cline{2-4}
& HOI synthesis &- &\cite{Ye2024G-HOP:Synthesis, Wang2024Single-ViewGeneration, Zhang2024ManiDext:Diffusion, Cao2024Multi-ModalGeneration, Li2024ClickDiff:Models, Zhang2024NL2Contact:Model, Lu2023Ugg:Grasping}\\
\cline{2-4}
& Object pose diffusion for reorientation and rearrangement &- & \cite{Liu2023StructDiffusion:Objects, Simeonov2023ShelvingRearrangement, Mishra2024ReorientDiff:Manipulation, Zhao2025AnyPlace:Manipulation}\\
\hline
\end{tabular}
\caption{\small Taxonomy of Grasp Generation Approaches with Diffusion Models}
    \label{tab:grasp-plan}
\end{table}

%
Grasp learning, as one of the crucial skills for robotic manipulation, has been studied over decades \citep{newbury2023deep}. Starting from hand-crafted feature engineering to statistical approaches \citep{bohg2013data}, accompanied by the recent progress in deep neural networks that are powered by massive data collection either from real-world 
\citep{fang2020graspnet} or simulated environments \citep{gilles2023metagraspnetv2, 10892643, shi2024vmf}. The current trend in grasp learning incorporates semantic-level object detection, leveraging open-vocabulary foundation models \citep{radford2021learning, liu2023grounding}, and focuses on object-centric or affordance-based grasp detection in the wild \citep{qian2024thinkgrasp, shi2025viso}. To this end, DMs, known for their ability to model complex distributions, allow for the creation of diverse and realistic grasp scenarios by simulating possible interactions with objects in a variety of contexts \citep{rombach2022high}. Furthermore, these models contribute to direct grasp generation by optimizing the generation of feasible and efficient grasps \citep{Urain2023SE3-DiffusionFields:Diffusion}, particularly in environments where real-time decision-making and adaptability are critical.

Grasp generation with DMs can be categorized into several key approaches: From methodological perspective, one category focuses on explicit diffusion on 6-DoF grasp poses that lie on the  \(\mathbf{SE}(3)\) group, directly modeling spatial transformations to generate feasible grasps \citep{Urain2023SE3-DiffusionFields:Diffusion, Song2024ImplicitGrasping, Wu2024LearningGrasping, Weng2024DexDiffuser:Models, singh2024constrained, lim2024equigraspflow}. Another line of approaches involves implicit grasp diffusion within latent space, enhancing adaptability and versatility \citep{Barad2023GraspLDM:Models}. A recent trend focuses on language-guided diffusion for task-oriented grasp generation,  where natural language inputs shape the generation process \citep{Nguyen2024LightweightModel, Vuong2024Language-drivenDetection, Nguyen2024Language-ConditionedClouds, Chang2024Text2Grasp:Parts}. Other approaches emphasize affordance-driven diffusion, targeting specific functional goals, such as object pose diffusion for rearrangement \citep{Liu2023StructDiffusion:Objects, Zhao2025AnyPlace:Manipulation}, affordance-guided object reorientation \citep{Mishra2024ReorientDiff:Manipulation}, imitation learning \citep{Wu2024Unidexfpm:Policy,Ma2024DexDiff:Environments} or multi-embodiment grasping \citep{freiberg2024diffusion}. Apart from these categories, hand-object interaction (HOI) specifically prioritizes the synthesis of realistic, functional interactions by modeling the hand's adaptive responses to various object shapes and affordances with dexterity  \citep{Ye2024G-HOP:Synthesis, Wang2024Single-ViewGeneration, Zhang2024ManiDext:Diffusion, Cao2024Multi-ModalGeneration, Li2024ClickDiff:Models, Zhang2024NL2Contact:Model, Lu2023Ugg:Grasping, zhang2024dexgraspnet}. In addition to the diffusion on grasp generation or trajectory planning, DM as sim-to-real generator \citep{Li2024ALDM-Grasping:Grasping} or foundational feature extractor \citep{Tsagkas2024ClickDescriptors} such as stable diffusion \citep{Rombach2021High-ResolutionModels} may provide semantic information to enhance downstream grasp generation tasks. \cref{tab:grasp-plan} summarizes the aforementioned categories. Notably, we include the applications of diffusion in HOI, imitation learning for pre-grasp, and tasks related to image generation in the graph, which will not be further discussed in the rest of this survey due to their relevance to the field of computer vision. While readers are still encouraged to refer to the relevant literature according to our illustration (\cref{tab:grasp-plan}: HOI Synthesis). More details on the architectures of the individual methods in grasp learning are provided in \cref{tab:benchmarks-grasps}.

\begin{table}[t]
\centering
\renewcommand{\arraystretch}{1.3}
\setlength{\tabcolsep}{4pt}
\small

\begin{tabular}{|l|l|l|l|l|}
\hline
\textbf{Reference} &\textbf{Input} & \textbf{Encoder} & \textbf{Diffuser} & \textbf{Benchmark} \\
\hline
\cite{Urain2023SE3-DiffusionFields:Diffusion} & SDF & Shape encoder & FiLM  & Acronym \\
\arrayrulecolor{lightgray}\hline\arrayrulecolor{black}
\cite{Barad2023GraspLDM:Models} & PCs & PointNet++ & FiLM & Acronym \\
\arrayrulecolor{lightgray}\hline\arrayrulecolor{black}
\cite{song2024implicit} & TSDF & OccNet & FiLM & VGN \\
\arrayrulecolor{lightgray}\hline\arrayrulecolor{black}
\cite{singh2024constrained} & PCs & OccNet & FiLM  & DA$^2$ \\
\arrayrulecolor{lightgray}\hline\arrayrulecolor{black}
\cite{lim2024equigraspflow} & PCs & VN-DGCNN & FiLM  & Acronym \\
\arrayrulecolor{lightgray}\hline\arrayrulecolor{black}
\cite{freiberg2024diffusion} & PCs + Gripper PCs & Eq. U-Net & Eq. FiLM  & Self generated \\
\arrayrulecolor{lightgray}\hline\arrayrulecolor{black}
\cite{carvalho2024grasp} & PCs & PointNet++ & DiT  & Acronym \\
\arrayrulecolor{lightgray}\hline\arrayrulecolor{black}
\cite{huang2025hgdiffuser} & PCs + Guidance & VN-PointNet & DiTs  & OakInk \\
\arrayrulecolor{lightgray}\hline\arrayrulecolor{black}
\cite{Weng2024DexDiffuser:Models} & PCs + Gripper PCs & BPS & DiTs  & DexGraspNet \\
\arrayrulecolor{lightgray}\hline\arrayrulecolor{black}
\cite{zhong2025gagrasp} & PCs + Gripper PCs & Eq. Models & Eq. DiTs  & MultiDex \\
\arrayrulecolor{lightgray}\hline\arrayrulecolor{black}
\cite{zhang2024dexgrasp} & PCs & PointNet++ & DiTs  & MultiDex \\
\hline
\end{tabular}
\caption{\small Technical details of grasp diffusion methodologies on \(\mathbf{SE}(3)\) grasp synthesis. The references for the encoders are provided in \cref{tab:ref-encoder}. The references for the benchmarks are listed in \cref{tab:benchmark-ref}. In the following, the abbreviations used are explained:
SDF:  Signed Distance Function, TSDF:  Truncated SDF, PCs:  Point Clouds, FiLM:  Convolutional Neural Network with Feature-wise Linear Modulation \cite{Perez2018FiLM:Layer}, DiTs:  Diffusion Transformers, Eq.:  Equivariant, VN:  Vector Neuron.}
\label{tab:benchmarks-grasps}
\end{table}

\subsubsection{Diffusion as \texorpdfstring{\(\mathbf{SE}(3)\)}{SE(3)} grasp pose generation}  
Since the standard diffusion process is primarily formulated in Euclidean space, directly extending it to \(\mathbf{SE}(3)\) poses, represented by:
\(
\mathbf{H} = 
\begin{bmatrix}
\mathbf{R} & \mathbf{t} \\
\mathbf{0} & 1
\end{bmatrix}
\)
is inherently challenging due to potential numerical instability (to satisfy \(\mathbf{H}\mathbf{H}^{-1} = \mathbf{I}^{4\times4}\)), since typical Langevin dynamics cannot be applied for non-Euclidean manifolds such as the \(\mathbf{SE}(3)\) Lie group. Here, \(\mathbf{R} \in \mathbf{SO}(3)\) represents the rotation matrix and \(\mathbf{t} \in \mathbb{R}^3\) the translation vector. Applying diffusion to \(\mathbf{SE}(3)\) poses requires accounting for the manifold's non-Euclidean nature, where standard Gaussian noise, as used in vanilla diffusion, fails to retain stability over rotations and translations.

To tackle this, \(\mathbf{SE}(3)\)-Diff \citep{Urain2023SE3-DiffusionFields:Diffusion} introduced a smooth cost function to learn the grasp quality via the energy-based model (EBM), where the score matching for EBM is applied on the Lie group to bridge the gap between diffusion processes on the vector space \(\mathbb{R}^6\) and the \(\mathbf{SE}(3)\). In contrast,  \citep{Song2024ImplicitGrasping} condition the 6-Dof grasp poses on the grasp locations \(\mathbf{t}\) and corresponding volumetric features for grasp generation in clutter following GIGA framework \citep{jiang2021synergies}, without explicit consideration on the \(\mathbf{SE}(3)\) constraint. Moreover, one advantage of the EBM model in \(\mathbf{SE}(3)\)-Diff is the direct grasp quality evaluation and integration into the entire grasp motion planning and optimization. However, training EBM-based models demands extensive sampling and poses significant challenges for generalization. We noticed that flow matching \citep{lipman2022flow}  is employed in recent studies, such as EquiGraspFlow \citep{lim2024equigraspflow} and Grasp Diffusion Network \citep{carvalho2024grasp}, which use continuous normalizing flows (CNFs) as ODE solvers to learn angular (\(\mathbf{SO}(3)\)) and linear (\(\mathbf{R}(3)\)) velocities for denoising. This preserves the \(\mathbf{SE}(3)\)-equivariance conditioned on the input point cloud given the time schedule. In contrast to \(\mathbf{SE}(3)\)-Diff, which relies on additional supervision in the form of signed distance functions, they achieve competitive performance without requiring this auxiliary module, leading to more efficient training. In general, although CNF-based approaches exhibit promising performance on grasp generation for a single object, more studies on generalizability to highly occluded environments \citep{freiberg2024diffusion} and uncertainty quantification \citep{shi2024vmf} are expected in future work. 

In contrast to explicit pose diffusion, latent DMs for grasp generation (GraspLDM \citep{Barad2023GraspLDM:Models}) explore latent space diffusion with VAEs, which does not explicitly account for the \(\mathbf{SE}(3)\) constraint. They follow VAE-based 6-Dof Graspnet \citep{Mousavian_2019_ICCV} to model the distribution of grasp latent features by a denoising diffusion process, which is conditioned on the point cloud and task latent for the grasp generation. This implicit modeling may potentially limit the model’s ability to generate physically plausible and geometrically consistent grasp poses.

Furthermore, the \(\mathbf{SE}(3)\) bi-equivariance property is critical for efficient grasp generation \citep{huang2023edge}, as it requires that any transformation applied to the input space correspondingly transforms the output space in a consistent manner. Specifically, this property implies that the generated poses from a \(\mathbf{SE}(3)\)-invariant distribution should maintain the same spatial and geometric relationships under transformations over the time schedule, ensuring that the learned grasp distribution remains invariant across various orientations and positions. For instance, Ryu et al.~\citep{ryu2022equivariant} consider bi-equivariance in Lie group representation to construct the equivariance descriptor field (EDF) \citep{ryu2022equivariant}, taking the transformations of both observation (target) space and initial end-effector frame into account. This principally improves the sample efficiency on pick-and-place tasks via Imitation learning. Upon this, they extend the EDF to bi-equivariant score matching \citep{Ryu2023Diffusion-EDFs:Manipulation} to be applied in the context of diffusion, which consists of both translational and rotational fields on \(\mathbf{se}(3)\) Lie algebra. Moreover, Freiberg et al. \citep{freiberg2024diffusion} adapts the approach from Ryu \citep{Ryu2023Diffusion-EDFs:Manipulation} to generalize to multi-embodiment grasping through an equivariant encoder that captures gripper embeddings. In terms of the theoretical background to equivariant robot learning, we identify a recent survey \citep{seo2025se} as a recommendation for interested readers.

\subsection{Visual data Augmentation}\label{chap:augmentation}
One line of methodologies focuses on employing mostly pretrained DMs for data augmentation in vision-based manipulation tasks. Here, the strong image generation and processing capabilities of diffusion generative models are utilized to augment data sets and scenes. The main goals of the visual data augmentation are scaling up data sets, scene reconstruction, and scene rearrangement.

\subsubsection{Scaling Data and Scene Augmentation}

A challenge associated with data-driven approaches in robotics relates to substantial data requirements, which are time-consuming to acquire, particularly for real-world data. In the domain of imitation learning, it is essential to accumulate an adequate number of expert demonstrations that accurately represent the task at hand. While, by now, many methods, e.g. \citep{Reuss2024MultimodalGoals,Ze20243DRepresentations,Ryu2023Diffusion-EDFs:Manipulation} only require a low number of five to fifty demonstrations, there are also methods, e.g. \citep{Chen2023PlayFusion:Play, Saha2024EDMP:Planning} relying on more extensive data sets. Especially offline RL methods, e.g. \citep{Carvalho2023MotionModels, Ajay2023ISDecision-Making} usually require extensive amounts of data to accurately predict actions over the complete state-action space, also from suboptimal behavior. Moreover, increasing the variability in training data also has the potential to increase the generalizability of the learned policies. 
Thus, to automatically increase the variety and size of datasets, without additional costs on researchers and staff, or other more engineering-heavy autonomous data collection pipelines \citep{Yu2023ScalingModels}, many methodologies, e.g. \citep{Chen2023GenAug:Environments, Mandi2022CACTI:Learning}, use DMs for data augmentation. In comparison to other strategies, such as domain randomization \citep{Trembley2018TrainingRandomization,Tobin2017Domainworld}, data augmentation with DMs directly augments the real-world data, making the data grounded in the physical world. In contrast, domain randomization requires complex tuning for each task, to ensure physical plausibility of the randomized scenes, and to enable sim-to-real transfer \citep{Chen2023GenAug:Environments}.


Given a set of real-world data, DM-based augmentation methods perform semantically meaningful augmentations via inpainting, such as changing object colors and textures \citep{Zhang2024DiffusionLearning}, or even replacing whole objects, as well as corresponding language task descriptions \citep{Chen2023GenAug:Environments, Yu2023ScalingModels, Mandi2022CACTI:Learning}. This enables both the augmentation of objects, which are part of the manipulation process, and backgrounds. The former increases the generalizability to different tasks and objects, while the latter increases robustness to scene information, which should not influence the policy. Some \citep{Zhang2024DiffusionLearning} also augment object positions and the corresponding trajectories to generate off-distribution demonstrations for DAgger, thus addressing the covariate shift problem in imitation learning. Others \citep{chen2024roviaug} augment camera view, robot embodiments, or even \citep{Katara2024Gen2Sim:Models} generate whole simulation scenes from given URDF files, prompted by a Large Language Model (LLM).
Targeted towards offline RL methods, \cite{DiPalo2024DiffusionLearning} combines data augmentation with a form of hindsight-experience replay \citep{NIPS2017_453fadbd} to adapt the visual observations to the language-task instruction. This increases the number of successful executions in the replay buffer, which potentially increases the data efficiency. The method is used to learn policies for new tasks, on previously collected data, to align the data with the new task instructions.


From a methodological perspective the methods mostly  employ frozen web-scale pretrained language \citep{Yu2023ScalingModels}, and vision-language models, for object segmentation \citep{Yu2023ScalingModels}, or text-to-image synthesis (Stable Diffusion) \citep{Rombach2021High-ResolutionModels, Mandi2022CACTI:Learning}, or  finetune \citep{Zhang2024DiffusionLearning, DiPalo2024DiffusionLearning} pretrained internet-scale vision-language models. 
Apart from \cite{Zhang2024DiffusionLearning} the methods, do not augment actions, but only observations. Thus, the methodologies must ensure augmentations, for which the demonstrated actions do not change, which highly limits the types of augmentations.
Moreover, large-scale data scaling via scene augmentation also requires additional computational cost. While this might not be a severe limitation, if it is applied once before the training, it may highly increase training time for online-RL methods.


\subsubsection{Sensor Data Reconstruction}

A challenge in vision-based robotic manipulation pertains to the incomplete sensor data. Especially single-view camera setups lead to incomplete object point clouds or images, making accurate grasp and trajectory prediction challenging. This is exacerbated by more complex task settings, with occlusion, as well as inaccurate sensor data.

Multiple methods \citep{Kasahara2023RIC:Reconstruction, 10802487} reconstruct camera viewpoints with DMs.
Given an RGBD image and camera intrinsics \cite{Kasahara2023RIC:Reconstruction} generates new object views without requiring CAD models of the objects. For this, the existing points are projected to the new viewpoint. The scene is segmented using the vision foundation model SAM \citep{Kirillov2023SegmentAnything}, to create object masks. On these masks missing data points are inpainted using the pretrained diffusion model for image generation Dall$\cdot$E \citep{Kapelyukh2023DALL-E-Bot:Robotics}. As Dall$\cdot$E does not ensure spatial consistency, consistency filtering is applied across viewpoints. Moreover, Dall$\cdot$E, only processes 2D images. Thus, to also complete the missing depth information, a model is trained to predict the missing depth information from the projected depth map and the reconstructed image.
In this method the viewpoints are sampled on evenly spaced directions along a viewing sphere. However, generating the point clouds for many viewpoints is computationally expensive, and might not be necessary for successful task completion. Thus, view-planning is applied to generate a minimal set of views. 
\citep{Pan2024ExploitingPlanning, pan2025dmosvp} use a DM to generate geometric priors from a 2D image, enabling a view-planner to sample a minimum set of viewpoints that minimize movement cost. The views are then used to train a Neural Radiance Field (NeRF) \citep{Mildenhall2020NeRF:Synthesis} to reconstruct 3D scenes from 2D images.

In the field of robotic manipulation, not many methods consider scene reconstruction. A possible reason for this is its relatively high computational cost. However, expanding to the areas of robotics and computer vision, more methodologies in the field of scene reconstruction exist. In robotic manipulation instead more methods focus on making policies more robust to incomplete or noisy sensor information, e.g. \citep{Ze20243DRepresentations, Ke20243DRepresentations}. However, the limited number of occlusion in the experimental setups indicate that strong occlusion are still a major challenge. Moreover, scene reconstruction is unable to react to completely occluded objects.

\subsubsection{Object Rearrangement}
The ability of DMs for text-to-image synthesis offers the possibility to generate plans from high-level task descriptions. In particular, given an initial visual observation, one group of methods uses such models to generate target-arrangement of objects in the scene, specified by a language-prompt\citep{Liu2023StructDiffusion:Objects, Kapelyukh2023DALL-E-Bot:Robotics, Xu2024SetModels, Zeng2024LVDiffusor:Diffusor, Kapelyukh2024Dream2Real:Models}. Examples of applications could be setting up a dinner table or clearing up a kitchen counter. While the earlier methodologies \citep{Kapelyukh2023DALL-E-Bot:Robotics, Liu2023StructDiffusion:Objects} use the pretrained VLM Dall$\cdot$E \citep{Black2024Zero-ShotModels} to generate rearrangements in a zero-shot manner, this has the disadvantage of possibly introducing scene inconsistencies and incompatibilities, due to the lack of geometric understanding and object permanence. Thus, the later methods \citep{Xu2024SetModels, Kapelyukh2024Dream2Real:Models} use combinations of pretrained LLMs and VLMs like CLIP \citep{Meila2021LearningSupervision}, together with other non-diffusion visual processing methods like NeRF \citep{Mildenhall2020NeRF:Synthesis} and SAM \citep{Kirillov2023SegmentAnything}, and custom DMs. The described methodologies are similar to the methods for object pose diffusion \citep{Mishra2024ReorientDiff:Manipulation, Simeonov2023ShelvingRearrangement, Zhao2025AnyPlace:Manipulation} mentioned in \cref{chap:grasp-learning}. The main difference is that the methods here focus on the rearrangement of multiple objects specified by a sparse language input, not exhaustively describing the geometric layout of the target arrangement. Different to the methods from \cref{chap:grasp-learning}, the integration with grasp or motion planning to achieve the target arrangement is not the focus.  However, nonetheless for all of the above listed methodologies for object rearrangement their effectiveness is also demonstrated in real-robot experiments.

\section{Experiments and Benchmarks}\label{chap:benchmarks}

In this section, we focus on the evaluation of the various DMs for robotic manipulation. Details on the employed benchmarks and baselines are listed in the separate tables for imitation learning (\cref{tab:benchmarks-il}), reinforcement learning (\cref{tab:benchmarks-rl}) in the \nameref{appendix}, and grasp learning (\cref{tab:benchmarks-grasps}). Separately, the references for all applied benchmarks are listed in \cref{tab:benchmark-ref} in the\nameref{appendix}.

Various benchmarks are used to evaluate the methods. Common benchmarks are CALVIN \citep{Mees2022CALVIN:Tasks}, RLBench \citep{James2020RLBench:Environment}, RelayKitchen \citep{Gupta2020RelayLearning}, and Meta-World \citep{Yu2019Meta-World:Learning}. Primarily in RL, the benchmark D4RL Kitchen \citep{Fu2020D4RL:Learning} is used. One method \citep{Ren2024DiffusionOptimization} uses FurnitureBench \citep{Heo2023FurnitureBench:Manipulation} for real-world manipulation tasks. Adroit \citep{Rajeswaran2017LearningDemonstrations} is a common benchmark for dexterous manipulation, LIBERO \citep{Liu2023LIBERO:Learning} for lifelong learning, and LapGym \citep{MariaScheikl2023LapGym-AnSurgery} for medical tasks. 

Many methods are only being evaluated against baselines, which are not based on DMs themselves. However, there are some common DM-based baselines. For methods operating in \(\mathbf{SE}(3)\)-space \citep{Chen2024DontDiffusion, Song2024ImplicitGrasping, Ryu2023Diffusion-EDFs:Manipulation}, \(\mathbf{SE}(3)\)-Diffusion Policy \citep{Urain2023SE3-DiffusionFields:Diffusion}, probably the first paper using DMs for grasp generation, is commonly used as baseline. For RL-based methods, the RL-based Diffuser \citep{Janner2022PlanningSynthesis}, Diffusion-QL \citep{Wang2023DiffusionLearning}, and Decision Diffuser \citep{Ajay2023ISDecision-Making} are commonly used as baselines. It should be noted that in the original paper, Decision Diffuser \citep{Ajay2023ISDecision-Making} is evaluated against Diffuser \citep{Janner2022PlanningSynthesis} and outperforms it on almost all tasks, particularly on the manipulation tasks, block stacking, and rearrangement. However, neither of these methods is evaluated on real-world tasks. 
Another common baseline is DP \citep{Chi2023DiffusionDiffusion}, as many methods are developed based on it. A common baseline for methods integrating 3D visual representations is 3D Diffusion Policy\citep{Ze20243DRepresentations}. 3D Diffusion Policy is evaluated against DP, and outperforms it on a huge variety of tasks in the benchmarks Adroit, MetaWorld, and Dexart with an average success rate of 74.4\%, outperforming DP by 24.2\%.  It is also evaluated on four real-world manipulation tasks: rolling and pinching a dumpling, drilling, and pouring. With an average success rate of 85.0\% it outperforms DP by 50\%. 
3D Diffusion Policy is greatly outperformed by 3D Diffuser Actor \citep{Ke20243DRepresentations} on the CALVIN benchmark, especially for zero-shot long-horizon tasks. However, no comparison for real-world tasks is provided.

The majority of methods are evaluated in simulation as well as in real-world experiments. For real-world experiments, most policies are directly trained on real-world data. However, some are trained exclusively in simulation and applied in the real world in a zero shot \citep{Yu2023ScalingModels, Mishra2023GenerativeModels, Ren2024DiffusionOptimization, Liu2023StructDiffusion:Objects, Kapelyukh2024Dream2Real:Models, Liu2023ComposableManipulation}, utilizing domain randomization, or real-world scene reconstruction in simulation.
Few, predominately RL methods, are only evaluated in simulation \citep{Yang2023CompositionalSolvers, Power2023SamplingModels, Wang2023DiffusionLearning, Janner2022PlanningSynthesis, Pearce2023ImitatingModels, Wang2023ColdStates, Mendez-Mendez2023EmbodiedPlanning, Kim2024StitchingRL, Brehmer2023EDGI:Agents, Liang2023AdaptDiffuser:Planners, Zhou2024VariationalExperts, Mishra2024ReorientDiff:Manipulation, Ajay2023ISDecision-Making, Ding2023ConsistencyLearning, Zhang2024LanguageTasks}.

\section{Conclusion, Limitations and Outlook}\label{chap:conclusion}
Diffusion models (DMs) have emerged as state-of-the-art methods in robotic manipulation, offering exceptional ability in modeling multi-modal distributions, high training stability, and stability to high-dimensional input and output spaces. Several tasks, challenges, and limitations in the domain of robotic manipulation with DMs remain unsolved. A prevalent issue is the lack of generalizability. 
The slow inference time for DMs also remains a major bottleneck. 


\subsection{Limitations}
\subsubsection{Generalizability}
While a lot of methods demonstrate relatively good generalizability in terms of object types, lightning conditions, and task complexity, they still face limitations in this area. This prevalent limitation shared across almost all methodologies in robotic manipulation remains a crucial research question.

The majority of methods using DMs for trajectory generation rely on imitation learning, using mostly behavior cloning. Thus, they inherit the dependence on the quality and diversity of training data, making it difficult to handle out-of-distribution situations due to the covariate shift problem \citep{pmlr-v9-ross10a}. 
As most methodologies combining DMs with RL use offline RL, they still rely on existing data, mapping a sufficient amount of the state-action space, and are thus also unable to react to distribution shifts. Moreover, offline RL requires more careful fine-tuning than imitation learning to ensure training stability and prevent overfitting.
Still, the advantage of RL is that it can handle suboptimal behavior \cite{DBLP:journals/corr/abs-2005-01643}.

While data scaling offers improved generalizability, it typically demands large training datasets and substantial computational resources. One recent solution is to use pre-trained foundation models. Moreover, as the majority of current methods for data augmentation in DMs do not augment trajectories, e.g \citep{Yu2023ScalingModels, Mandi2022CACTI:Learning}, it only increases robustness to slightly different task settings, such as changes in colors, textures, distractors, and background.
VLAs can generalize to multi-task and long-horizon settings but often lack action precision, thus requiring finetuning and the combination with more specialized agents \citep{zhang2024grapegeneralizingrobotpolicy}.

\subsubsection{Sampling speed}
The principal limitation inherent to DMs can be attributed to the iterative nature of the sampling process, which results in a time-intensive sampling procedure, thus impeding efficiency and real-time prediction capabilities. Despite recent advances that improve sampling speed and quality \citep{Chen2024DontDiffusion, Zhou2024VariationalExperts}, a considerable number of recent methods use DDIM \citep{Song2021DenoisingModels}, although other methods, such as DPM-solver \citep{Lu2022DPM-Solver:Steps} have shown better performance. However, this comparison has only been performed using image generation benchmarks and would need to be verified for applications in robotic manipulation. Numerous works have demonstrated competitive task performance using DDIM, but do not directly investigate the decrease in task performance associated with a lower number of reverse diffusion steps. \cite{Ko2024LearningCorrespondences} analyzes their approach using both DDPM and DDIM sampling, reporting a sampling process that is ten times faster with only a 5.6\% decrease in task performance when using DDIM. Although such a decline might appear negligible, its significance is highly task-dependent. Consequently, there is a need for efficient sampling strategies and a more comprehensive analysis of existing sampling methods, particularly regarding the domain of robotic manipulation. It should, however, be noted that already in DP \citep{Chi2023DiffusionDiffusion}, one of the earlier methods combining DMs with receding-horizon control for trajectory planning, real-time control is possible. Using DDIM with 10 denoising steps during inference, they report an inference latency of 0.1s on a Nvidia 3080 GPU.

\subsection{Conclusion and Outlook}
This survey, to the best to our knowledge, is the first survey reviewing the
state-of-the-art methods diffusion models (DMs) in robotics manipulation. This paper offers a thorough discussion of various methodologies regarding network architecture, learning framework, application, and evaluation, highlighting limits and advantages.  We explored the three primary applications of DMs in robotic manipulation: trajectory generation, robotic grasping, and visual data augmentation. 
Most notably, DMs offer exceptional ability in modeling multi-modal distributions, high training stability, and robustness to high-dimensional input and output spaces. Especially in visual robotic manipulation, DMs provide essential capabilities to process high-resolution 2D and 3D visual observations, as well as to predict high-dimensional trajectories and grasp poses, even directly in image space.

A key challenge of DMs is the slow inference speed.
In the field of computer vision, fast samplers have been developed that have not yet been evaluated in the field of robotic manipulation.
Testing those samplers and comparing them against the commonly used ones, could be one step to increase sampling efficiency. Moreover, there are also methods for fast sampling, specifically in robotic manipulation, that are not broadly used, e.g. BRIDGeR \citep{Chen2024DontDiffusion}. 
While the generalizability of DMs remains also an open challenge, the image generation capabilities of DMs open new avenues in data augmentation for data scaling, making methods more robust to limited data variety. Generalizability could be also improved by the integration of advanced vision-language, and vision-language action models. 

We believe continual learning could be a promising approach to improve generalizability and adaptability in highly dynamic and unfamiliar environments. This remains
a widely unexplored problem domain for DMs in robotic manipulation, exceptions are \citep{DiPalo2024DiffusionLearning, Mendez-Mendez2023EmbodiedPlanning}. However, these methods have strong limitations. For instance, \citep{DiPalo2024DiffusionLearning} relies on precise feature descriptions of all involved objects and is restricted to predefined abstract skills. Moreover, their continual update process involves replaying all past data, which is both computationally inefficient and does not prevent catastrophic forgetting.
Morover, to handle complex and cluttered scenes, view planning and iterative planning strategies, also considering complete occlusions, could be combined with existing DMs using 3D scene representations. Leveraging the semantic reasoning capabilities of vision language and vision language action models could be a possible approach.

\section*{Acknowledgment}
We thank our colleague Edgar Welte for providing the video data for the illustration of the diffusion process in \cref{fig:fig2}.

\section*{Funding}
Funded by the Deutsche Forschungsgemeinschaft (DFG, German
Research Foundation) – SFB-1574 – 471687386

\bibliographystyle{apalike}
\bibliography{references}

\newpage
\section*{Appendix}\label{appendix}

\begin{table}[h!]
\centering
\setlength{\tabcolsep}{4pt}
\small
\begin{tabular}{|L{3.7cm}|L{3.2cm}|L{3cm}|L{1.8cm}|C{1cm}|L{1.5cm}|}
\hline
\textbf{Reference} & \textbf{Diffusion Baseline} & \multicolumn{2}{|c|}{\textbf{Simulation}} & \multicolumn{2}{|c|}{\textbf{Real World}}\\
\cline{3-6}
  &  & \textbf{Benchmark} & \textbf{\#Demos} & \textbf{Real} & \textbf{\#Demos} \\
\hline
Diffusion Policy (DP) \citep{Chi2023DiffusionDiffusion} & \bxmark & FrankaKitchen, Robomimic, custom & 566,\newline 500,\newline / & \bcmark & / \\
\arrayrulecolor{lightgray}\hline\arrayrulecolor{black}
ChainedDiffuser \citep{Xian2023ChainedDiffuser:Manipulation} & \bxmark & RLBench & 100 & \bcmark & 10 - 20 \\
\arrayrulecolor{lightgray}\hline\arrayrulecolor{black}
BESO \citep{Reuss2023Goal-ConditionedPolicies} & DP, Diffusion-BC & Relay Kitchen, CALVIN, custom & 566,\quad\quad\quad /, 1000 & \bxmark & - \\
\arrayrulecolor{lightgray}\hline\arrayrulecolor{black}
\cite{Chen2023PlayFusion:Play} & \bxmark & CALVIN, FrankaKitchen, Ravens & 200K, \newline566, \newline1000 & \bcmark & / \\
\arrayrulecolor{lightgray}\hline\arrayrulecolor{black}
\cite{Zhou2023AdaptiveModels} & Diffuser$^{*1}$,\newline Decision Diffuser$^{*2}$ & RLBench &  & \bxmark & - \\
\arrayrulecolor{lightgray}\hline\arrayrulecolor{black}
Diffusion-BC \citep{Pearce2023ImitatingModels} & \bxmark & D4RLKitchen & 566 & \bxmark & - \\
\arrayrulecolor{lightgray}\hline\arrayrulecolor{black}
\cite{Mendez-Mendez2023EmbodiedPlanning} & \bxmark & BEHAVIOR & - & \bxmark & - \\
\arrayrulecolor{lightgray}\hline\arrayrulecolor{black}
3D-DP \citep{Ze20243DRepresentations} & DP & e.g. Adroit, MetaWorld, DexDeform & 10 - 100 & \bcmark & 40 \\
\arrayrulecolor{lightgray}\hline\arrayrulecolor{black}
\cite{Ke20243DRepresentations} & 3D-DP, ChainedDiffuser & RLBench, CALVIN & 24h & \bcmark & 15 \\
\arrayrulecolor{lightgray}\hline\arrayrulecolor{black}
\cite{Liu2023ComposableManipulation} & DP, SE(3)-DM & custom & / & \bcmark & / \\
\arrayrulecolor{lightgray}\hline\arrayrulecolor{black}
\cite{Power2023SamplingModels} & \bxmark & custom & / & \bxmark & - \\
\arrayrulecolor{lightgray}\hline\arrayrulecolor{black}
\cite{Ma2024HierarchicalManipulation} & DP, Diffuser & RLBench & 100 & \bcmark & 20 \\
\arrayrulecolor{lightgray}\hline\arrayrulecolor{black}
\cite{Vosylius2024RenderCloning} & DP & RLBench & 20 & \bcmark & 20 \\
\arrayrulecolor{lightgray}\hline\arrayrulecolor{black}
\cite{Zhang2024LanguageTasks} & Diffuser & CALVIN & / & \bxmark & - \\
\arrayrulecolor{lightgray}\hline\arrayrulecolor{black}
\cite{Reuss2024MultimodalGoals} & \bxmark & CALVIN, LIBERO & 24h, 50 & \bcmark & 4.5h \\
\arrayrulecolor{lightgray}\hline\arrayrulecolor{black}
\cite{Scheikl2024MovementObjects} & DP, BESO & LapGym & 90 - 200 & \bcmark & 90 - 200 \\
\arrayrulecolor{lightgray}\hline\arrayrulecolor{black}
\cite{Chen2024DontDiffusion} & DP, SE(3)-DM & FrankaKitchen, Adroit & 16k - 64k, 1.25k - 5k & \bcmark & 60 \\
\arrayrulecolor{lightgray}\hline\arrayrulecolor{black}
\cite{Zhou2024VariationalExperts} & DP, BESO, Consistency Models$^{*3}$ & Relay Kitchen, XArm Block Push, D3IL & 566,\newline 1k,\newline 96 - 2k & \bxmark & - \\
\arrayrulecolor{lightgray}\hline\arrayrulecolor{black}
\cite{Li2024CrosswayLearning} & DP, 3D-DP & Robomimic, custom & 500,\newline 100 & \bcmark & 100 \\
\arrayrulecolor{lightgray}\hline\arrayrulecolor{black}
\cite{Si2024Tilde:DeltaHand} & DP & \bxmark & - & \bcmark & 25 - 50 \\
\arrayrulecolor{lightgray}\hline\arrayrulecolor{black}
\cite{Saha2024EDMP:Planning} & \bxmark & M$\pi$Nets & 6.54Mil & \bxmark & - \\
\arrayrulecolor{lightgray}\hline\arrayrulecolor{black}
\cite{Bharadhwaj2024Track2Act:Manipulation} & \bxmark & EpicKitchens, RT1, BridgeData & 400k $^{*4}$ & \bcmark & 400 \\
\arrayrulecolor{lightgray}\hline\arrayrulecolor{black}
\cite{Wang2024PoCo:Learning} & \bxmark & custom & 50K trans$^{*5}$ & \bcmark & 50K trans$^{*5}$ \\
\arrayrulecolor{lightgray}\hline\arrayrulecolor{black}
\cite{Li2024GeneralizableGuidance} & DP, 3D-DP & RLBench & 40 & \bcmark & 40 \\
\arrayrulecolor{lightgray}\hline\arrayrulecolor{black}
\cite{reuss2024efficient} & DP & CALVIN, LIBERO, Relay Kitchen, \newline Block Push & 22966, 50, 566, \newline1000 & \bxmark & -  \\

\hline
\end{tabular}
\caption{\small Benchmarks of trajectory diffusion using imitation learning. For each benchmark, the numbers of demonstrations are listed in the same order. In the column ``Diffusion Baselines'' only those baselines, which are diffusion methods themselves, are listed. Methods not evaluated against a diffusion-based baseline, indicated by an (\bxmark), are only evaluated against non-diffusion baselines or ablations of the method.\\
The references for the benchmarks are listed in \cref{tab:benchmark-ref}. In the following, the symbols are explained: Methods by $^{*1}$\cite{Janner2022PlanningSynthesis}, $^{*2}$ \cite{Ajay2023ISDecision-Making}, and  $^{*3}$ \citep{song2023consistencymodels}. $^{*4}$ The diffusion model is trained using uncurated video data. $^{*5}$ As the number refers to the number of transitions, not demonstrations, this high number is expected.
The column ``Real'' indicates whether methods are evaluated in the real world (\bcmark), or not (\bxmark).
A~``/'' indicates that the information is not provided by the cited paper, while a ``-'' indicates that the information does not apply.
}
\label{tab:benchmarks-il}
\end{table}

\setlength{\tabcolsep}{8pt}

\begin{table}[h!]
\centering
\setlength{\tabcolsep}{4pt}
\small
\begin{tabular}{|L{3.7cm}|L{3.2cm}|L{3cm}|L{2.1cm}|C{1cm}|L{1.5cm}|}
\hline
\textbf{Reference} & \textbf{Diffusion Baseline} & \multicolumn{2}{|c|}{\textbf{Simulation}} & \multicolumn{2}{|c|}{\textbf{Real World}}\\
\cline{3-6}
  &  & \textbf{Benchmark} & \textbf{\#Demos} & \textbf{Real} & \textbf{\#Demos} \\
\hline
Diffuser \citep{Janner2022PlanningSynthesis} & \bxmark \color{black} & KUKA (custom) & 10k & \bxmark & - \\
\arrayrulecolor{lightgray}\hline\arrayrulecolor{black}
Decision Diffuser \citep{Ajay2023ISDecision-Making} & \bxmark \color{black} & D4RLKitchen, KUKA & /, 10k & \bxmark & - \\
\arrayrulecolor{lightgray}\hline\arrayrulecolor{black}
Diffusion-QL \citep{Wang2023DiffusionLearning} & \bxmark & D4RLKitchen & 10000 trans$^{*1}$ & \bxmark & - \\
\arrayrulecolor{lightgray}\hline\arrayrulecolor{black}
\cite{Wang2023ColdStates} & Diffuser & custom & 8k & \bxmark & - \\
\arrayrulecolor{lightgray}\hline\arrayrulecolor{black}
HDMI \citep{pmlr-v202-li23ad} & \bxmark & \bcmark & - & \bxmark & - \\
\arrayrulecolor{lightgray}\hline\arrayrulecolor{black}
\cite{Ding2023ConsistencyLearning} & Diffusion-QL & D4RLKitchen, Adroit & / & \bxmark & - \\
\arrayrulecolor{lightgray}\hline\arrayrulecolor{black}
\cite{Mishra2023GenerativeModels} & Decision Diffuser & STAP & / & \bcmark & / \\
\arrayrulecolor{lightgray}\hline\arrayrulecolor{black}
\cite{Kang2023EfficientLearning} & Diffusion-QL & Adroit, D4RL Kitchen & / & \bxmark & - \\
\arrayrulecolor{lightgray}\hline\arrayrulecolor{black}
\cite{Brehmer2023EDGI:Agents} & Diffuser & KUKA & / & \bxmark & - \\
\arrayrulecolor{lightgray}\hline\arrayrulecolor{black}
\cite{Suh2023FightingMatching} & Diffuser & \bcmark & - & \bcmark & 100 \\
\arrayrulecolor{lightgray}\hline\arrayrulecolor{black}
\cite{Ha2023ScalingAcquisition} & Diffusion-QL & \bxmark & - & \bcmark & 50 \\
\arrayrulecolor{lightgray}\hline\arrayrulecolor{black}
\cite{Kim2024StitchingRL} & Diffuser, Decision Diffuser, HDMI & Fetch env  & / & \bxmark & - \\
\arrayrulecolor{lightgray}\hline\arrayrulecolor{black}
\cite{Liang2023AdaptDiffuser:Planners} & Diffuser, Decision Diffuser & KUKA & / & \bxmark & - \\
\arrayrulecolor{lightgray}\hline\arrayrulecolor{black}
\cite{Zhang2024LanguageTasks} & Diffuser & CALVIN, CLEVR-Robot & / & \bxmark & - \\
\arrayrulecolor{lightgray}\hline\arrayrulecolor{black}
\cite{Ada2024DiffusionLearning} & Diffusion-QL & \bcmark & / & \bcmark & - \\
\arrayrulecolor{lightgray}\hline\arrayrulecolor{black}
\cite{Ren2024DiffusionOptimization} & Diffusion-QL & Robomimic, \newline D3IL,\newline FurnitureBench & 100-300, \newline96,\newline 50 & \bcmark \color{black}$^{*2}$ & 50 \\
\arrayrulecolor{lightgray}\hline\arrayrulecolor{black}
\cite{Huang2025DiffusionDiffusion} & Diffusion-QL & MetaWorld, Adroit & 20, 50 & \bcmark & 50 \\
\arrayrulecolor{lightgray}\hline\arrayrulecolor{black}
\cite{Carvalho2023MotionModels} & \bxmark & custom & 25 & \bxmark & - \\

\hline
\end{tabular}
\caption{\small Benchmarks of trajectory diffusion using reinforcement learning. For each benchmark, the numbers of demonstrations are listed in the same order. In the column ``Diffusion Baselines'' only those baselines, which are diffusion methods themselves, are listed. Methods not evaluated against a diffusion-based baseline, indicated by an (\bxmark), are only evaluated against non-diffusion baselines or ablations of the method.
The references for the benchmarks are listed in \cref{tab:benchmark-ref}. In the following, the symbols are explained: $^{*1}$ As the number refers to the number of transitions, not demonstrations, this high number is expected.
A (\bcmark) in the column ``Benchmark'' indicates that the method is evaluated in simulation, but not with a robotic manipulation task, while a (\bxmark) indicates that the method is not evaluated in simulation.
The column ``Real'' indicates whether methods are evaluated in the real world (\bcmark), or not (\bxmark).
A~``/'' indicates that the information is not provided by the cited paper, while a ``-'' indicates that the information does not apply.}
\label{tab:benchmarks-rl}
\end{table}

\begin{table}[h]
\centering
\renewcommand{\arraystretch}{1.1}
\setlength{\tabcolsep}{4pt}
\small
\begin{tabular}{|l|l|}
\hline
\textbf{Encoder} & \textbf{Reference} \\
\hline
\rowcolor{white!80!black}
\multicolumn{2}{|l|}{\textbf{Vision}} \\
\hline
ResNet & \cite{He_2016_CVPR} \\
\arrayrulecolor{lightgray}\hline\arrayrulecolor{black}
PointNet++ & \cite{qi2017pointnet++}\\
\arrayrulecolor{lightgray}\hline\arrayrulecolor{black}
Vision Transformer (ViT) & \cite{dosovitskiy2020image}\\
\arrayrulecolor{lightgray}\hline\arrayrulecolor{black}
VQ-GAN & \cite{9578911}\\
\arrayrulecolor{lightgray}\hline\arrayrulecolor{black}
OccNet & \cite{mescheder2019occupancy}\\
\arrayrulecolor{lightgray}\hline\arrayrulecolor{black}
VN-DGCNN & \cite{deng2021vector}\\
\arrayrulecolor{lightgray}\hline\arrayrulecolor{black}
Equivariant U-Net & \cite{ryu2022equivariant}\\
\arrayrulecolor{lightgray}\hline\arrayrulecolor{black}
VN-PointNet & \cite{deng2021vector}\\
\arrayrulecolor{lightgray}\hline\arrayrulecolor{black}
BPS & \cite{prokudin2019efficient}\\
\arrayrulecolor{lightgray}\hline\arrayrulecolor{black}
ShapeEncoder & \cite{park2019deepsdf}\\
\hline
\rowcolor{white!80!black}
\multicolumn{2}{|l|}{\textbf{Vision-Language}} \\
\hline
CLIP & \cite{radford2021learning}\\
\arrayrulecolor{lightgray}\hline\arrayrulecolor{black}
SAM & \cite{Kirillov2023SegmentAnything}\\
\arrayrulecolor{lightgray}\hline\arrayrulecolor{black}
XMem & \cite{10.1007/978-3-031-19815-1_37}\\
\arrayrulecolor{lightgray}\hline\arrayrulecolor{black}
HULC & \cite{9849097}\\
\arrayrulecolor{lightgray}\hline\arrayrulecolor{black}
T5 & \cite{JMLR:v21:20-074}\\
\hline
\end{tabular}
\caption{\small References for architectures of encoders, for different input modalities.}
\label{tab:ref-encoder}
\end{table}
\setlength{\tabcolsep}{4pt}

\begin{table}[h]
\centering
\small
\begin{tabular}{|p{5.5cm}|p{6.5cm}|}
\hline
\textbf{Dataset} & \textbf{Reference} \\ 
\hline
\rowcolor{white!80!black}
\multicolumn{2}{|l|}{\textbf{Trajectories}} \\ 
\hline
Adroit & \cite{Fu2020D4RL:Learning} \\ \arrayrulecolor{lightgray}\hline\arrayrulecolor{black}
BEHAVIOR & \cite{pmlr-v164-srivastava22a} \\ \arrayrulecolor{lightgray}\hline\arrayrulecolor{black}
BridgeData & \cite{pmlr-v229-walke23a} \\ \arrayrulecolor{lightgray}\hline\arrayrulecolor{black}
CALVIN & \cite{Mees2022CALVIN:Tasks} \\ \arrayrulecolor{lightgray}\hline\arrayrulecolor{black}
D3IL & \cite{JiaBlessingJiang2024_1000175328} \\ \arrayrulecolor{lightgray}\hline\arrayrulecolor{black}
D4RLKitchen & \cite{Fu2020D4RL:Learning} \\ \arrayrulecolor{lightgray}\hline\arrayrulecolor{black}
DexDeform & \cite{Ma2024DexDiff:Environments} \\ \arrayrulecolor{lightgray}\hline\arrayrulecolor{black}
EpicKitchens & \cite{Damen2021PAMI} \\ \arrayrulecolor{lightgray}\hline\arrayrulecolor{black}
Fetch env & \cite{NEURIPS2022_022a3905} \\ \arrayrulecolor{lightgray}\hline\arrayrulecolor{black}
FrankaKitchen & \cite{Gupta2020RelayLearning} \\ \arrayrulecolor{lightgray}\hline\arrayrulecolor{black}
FurnitureBench & \cite{Heo2023FurnitureBench:Manipulation}\\ \arrayrulecolor{lightgray}\hline\arrayrulecolor{black}
KUKA & Diffuser \\ \arrayrulecolor{lightgray}\hline\arrayrulecolor{black}
LabGym & \cite{MariaScheikl2023LapGym-AnSurgery} \\ \arrayrulecolor{lightgray}\hline\arrayrulecolor{black}
LIBERO & \cite{Liu2023LIBERO:Learning} \\ \arrayrulecolor{lightgray}\hline\arrayrulecolor{black}
MetaWorld & \cite{Yu2019Meta-World:Learning} \\ \arrayrulecolor{lightgray}\hline\arrayrulecolor{black}
M$\pi$Nets & \cite{pmlr-v205-fishman23a} \\ \arrayrulecolor{lightgray}\hline\arrayrulecolor{black}
STAP & \cite{agia2022taps} \\ \arrayrulecolor{lightgray}\hline\arrayrulecolor{black}
Ravens & \cite{pmlr-v155-zeng21a, pmlr-v164-shridhar22a} \\ \arrayrulecolor{lightgray}\hline\arrayrulecolor{black}
Relay Kitchen & \cite{Gupta2020RelayLearning} \\ \arrayrulecolor{lightgray}\hline\arrayrulecolor{black}
RLBench & \cite{James2020RLBench:Environment} \\ \arrayrulecolor{lightgray}\hline\arrayrulecolor{black}
Robomimic & \cite{pmlr-v164-mandlekar22a} \\ \arrayrulecolor{lightgray}\hline\arrayrulecolor{black}
RT1 & \cite{brohan2023rt1roboticstransformerrealworld} \\ \arrayrulecolor{lightgray}\hline\arrayrulecolor{black}
XArm Block Push & \cite{Florence2022ImplicitCloning} \\ 
\hline\rowcolor{white!80!black}
\multicolumn{2}{|l|}{\textbf{Grasps}}\\ 
\hline
Acronym & \cite{eppner2021acronym}\\ \arrayrulecolor{lightgray}\hline\arrayrulecolor{black}
DA$^2$ & \cite{zhai20222}\\ \arrayrulecolor{lightgray}\hline\arrayrulecolor{black}
DexGraspNet & \cite{wang2023dexgraspnet}\\ \arrayrulecolor{lightgray}\hline\arrayrulecolor{black}
MultiDex & \cite{Li2024ClickDiff:Models}\\ \arrayrulecolor{lightgray}\hline\arrayrulecolor{black}
OakInk & \cite{yang2022oakink}\\ \arrayrulecolor{lightgray}\hline\arrayrulecolor{black}
VGN & \cite{breyer2021volumetric}\\ 
\hline

\end{tabular}
\caption{\small List of datasets and their corresponding references for trajectory diffusion and grasp diffusion.}
\label{tab:benchmark-ref}
\end{table}

\end{document}